\documentclass[sigconf,10pt]{acmart}

\setcopyright{rightsretained}

\usepackage{tikz}
\usepackage{amsmath}
\usepackage{amsthm}
\usepackage{colortbl}
\usepackage{graphicx}
\usepackage{epsfig,url,xcolor,xspace}
\usepackage{balance}
\usepackage{enumitem}
\usepackage{tabularx}
\usepackage{multirow}
\usepackage{listings}
\usepackage{anyfontsize}
\usepackage{hyperref}
\usepackage{nccmath}
\usepackage[ruled,linesnumbered]{algorithm2e}
\usepackage{subcaption}
\usepackage[T1]{fontenc}
\usepackage[scaled=0.85]{beramono}
\usepackage{flushend}

\hypersetup{
    colorlinks=true,
    citecolor=black,
    urlcolor=black, %
    linkcolor=black, %
}

\makeatletter
\newcommand\BeraMonottfamily{%
  \def\fvm@Scale{0.85}%
  \fontfamily{fvm}\selectfont%
}
\makeatother

\lstdefinestyle{codestyle}{
    language=Python,
    basicstyle=\small\BeraMonottfamily,
    keywordstyle=\bfseries,
    commentstyle=\color{green},
    numberstyle=\tiny\color{gray},
    belowskip=-1em,
    breakatwhitespace=false,
    breaklines=true,
    captionpos=b,
    keepspaces=true,
    numbers=none,
    numbersep=5pt,
    showspaces=false,
    showstringspaces=false,
    showtabs=false,
    tabsize=2,
    frame=lines,
    morekeywords={apply,batch,bucket\_by\_sequence\_length,distribute,flat\_map,from_tensors,group\_by\_window,interleave,map,prefetch,range,TextFileDataset}
}

\lstdefinestyle{Python}{
    language        = Python,
    basicstyle      = \small\ttfamily,
    keywordstyle    = \color{blue},
    emph            = {list\_files,interleave,map,batch,prefetch,distribute,bucket\_by\_sequence\_length,group\_by\_window,filter,flat\_map},
    emphstyle       = \color[rgb]{0.4376,0.6463,0.3471},
    keywords        = {[2]num\_parallel\_calls,batch\_size,processing\_mode,service\_address,service,job\_name,num\_consumers,consumer\_index,bucket\_boundaries,window\_size}, 
    keywordstyle    = {[2]\color{orange}},
    stringstyle     = \color{red}\ttfamily,
}

\definecolor{ETHa}{RGB}{168,50,45}      %
\definecolor{ETHb}{RGB}{115,191,15}     %
\definecolor{ETHc}{RGB}{18,105,176}     %
\definecolor{ETHd}{RGB}{191,15,135}     %
\definecolor{ETHe}{RGB}{145,5,106}      %
\definecolor{ETHf}{RGB}{252,186,3}      %
\definecolor{ETHg}{RGB}{31,64,122}      %
\definecolor{ETHh}{RGB}{0,122,150}      %
\definecolor{ETHi}{RGB}{149,96,19}      %

\newboolean{showcolor}   
\setboolean{showcolor}{false}
\newcommand{\review}[2]{%
    \ifthenelse{\boolean{showcolor}}{%
        \ifnum #1=1%
            {\color{ETHa} #2}
        \fi%
        \ifnum #1=2%
            {\color{ETHi} #2}
        \fi%
        \ifnum #1=3%
            {\color{ETHc} #2}
        \fi%
        \ifnum #1=4%
            {\color{ETHd} #2}
        \fi%
        \ifnum #1=5%
            {\color{ETHe} #2}
        \fi%
        \ifnum #1=6%
            {\color{ETHf} #2}
        \fi%
        \ifnum #1=7%
            {\color{ETHg} #2}
        \fi%
        \ifnum #1=8%
            {\color{ETHh} #2}
        \fi%
        \ifnum #1=9%
            {\color{ETHb} #2}
        \fi%
    }{%
        #2%
    }%
}

\newcommand{\ignore}[1]{}

\newcommand{\inline}[1]{\lstinline[style=Python]{#1}}

\newcommand{\tfdata}{tf.data\xspace}
\newcommand{\tfdataservice}{tf.data service\xspace}

\renewcommand{\paragraph}[1]{\vspace{3pt}\noindent\textbf{#1. }\xspace}

\theoremstyle{definition}

\acmYear{2023}\copyrightyear{2023}
\acmConference[SoCC '23]{ACM Symposium on Cloud Computing}{October
30--November 1, 2023}{Santa Cruz, CA, USA}
\acmBooktitle{ACM Symposium on Cloud Computing (SoCC '23), October
30--November 1, 2023, Santa Cruz, CA, USA}
\acmPrice{15.00}
\acmDOI{10.1145/3620678.3624666}
\acmISBN{979-8-4007-0387-4/23/11}

\begin{document}

\title[tf.data service: A Case for Disaggregating ML Input Data Processing]{tf.data service: A Case for Disaggregating ML Input Data Processing}

\author{Andrew Audibert}
\affiliation{%
  \institution{Google}
  \country{}
}

\author{Yang Chen}
\affiliation{%
  \institution{Google}
  \country{}
}

\author{Dan Graur}
\affiliation{%
  \institution{ETH Z\"urich}
  \country{}
}

\author{Ana Klimovic}
\affiliation{%
  \institution{ETH Z\"urich}
  \country{}
}

\author{Ji\v{r}\'{i} \v{S}im\v{s}a}
\affiliation{%
  \institution{Google}
  \country{}
}

\author{Chandramohan A. Thekkath}
\affiliation{%
  \institution{Google}
  \country{}
}

\renewcommand{\shortauthors}{Audibert A., Chen Y., Graur D., Klimovic A., \v{S}im\v{s}a J., and Thekkath C.}

\begin{abstract}

Machine learning (ML) computations commonly execute on expensive specialized hardware, such as GPUs and TPUs, which provide high FLOPs and performance-per-watt. For cost efficiency, it is essential to keep these accelerators highly utilized. This requires preprocessing input data at the rate at which the accelerators can ingest and perform ML computations on the data. To avoid data stalls, the host CPU and RAM required for input data processing per accelerator core used for ML computations varies across jobs. Hence, the traditional approach of processing input data on ML accelerator hosts with a fixed hardware ratio leads to either under-utilizing the accelerators or the host CPU and RAM. In this paper, we address these concerns by building a disaggregated ML data processing system.

We present \textit{tf.data service}, an open-source disaggregated input data processing service built on top of \tfdata in TensorFlow. We show that disaggregating data preprocessing has three key advantages for large-scale ML training jobs. First, the service can horizontally scale-out to right-size CPU/RAM host resources for data processing in each job, saving 32$\times$ training time and 26$\times$ cost, on average. Second, the service can share ephemeral preprocessed data results across jobs, to optimize CPU usage and reduce redundant computations. Finally, the service supports coordinated reads, a technique that avoids stragglers due to different input sizes in distributed training, reducing training time by 2.2$\times$, on average. Our design is inspired by lessons learned from deploying tf.data service in production, including relaxing data visitation guarantees without impacting model accuracy.

\end{abstract}

\maketitle

\section{Introduction}
\label{sec:intro}

\begin{figure*}[htb]
  \centering
  \begin{subfigure}[t]{0.32\textwidth}
    \centering
    \includegraphics[width=\textwidth]{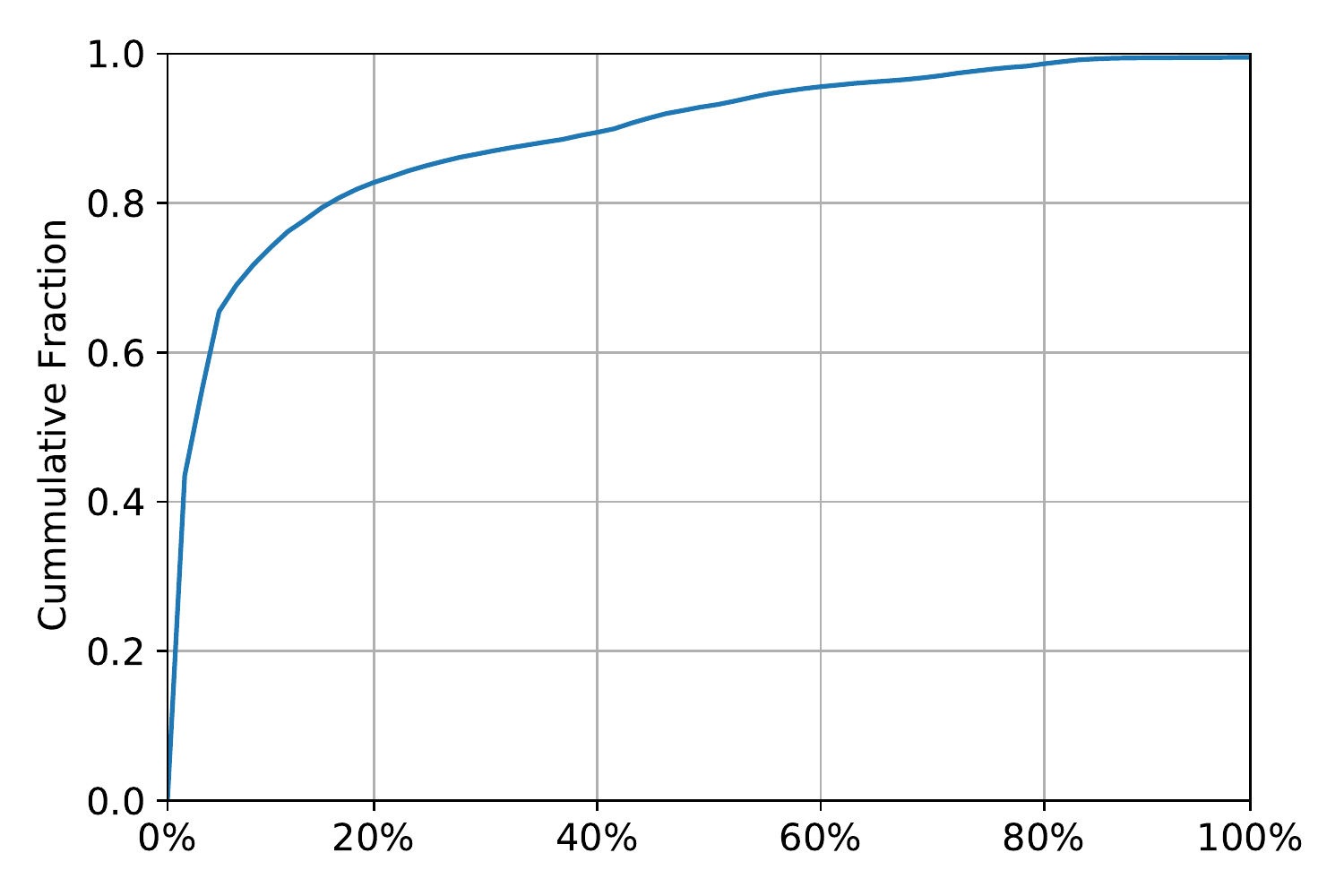}
    \caption{CPU usage}
    \label{fig:intro:cpu-analysis}
  \end{subfigure}
  \hfill
  \begin{subfigure}[t]{0.32\textwidth}
    \centering
    \includegraphics[width=\textwidth]{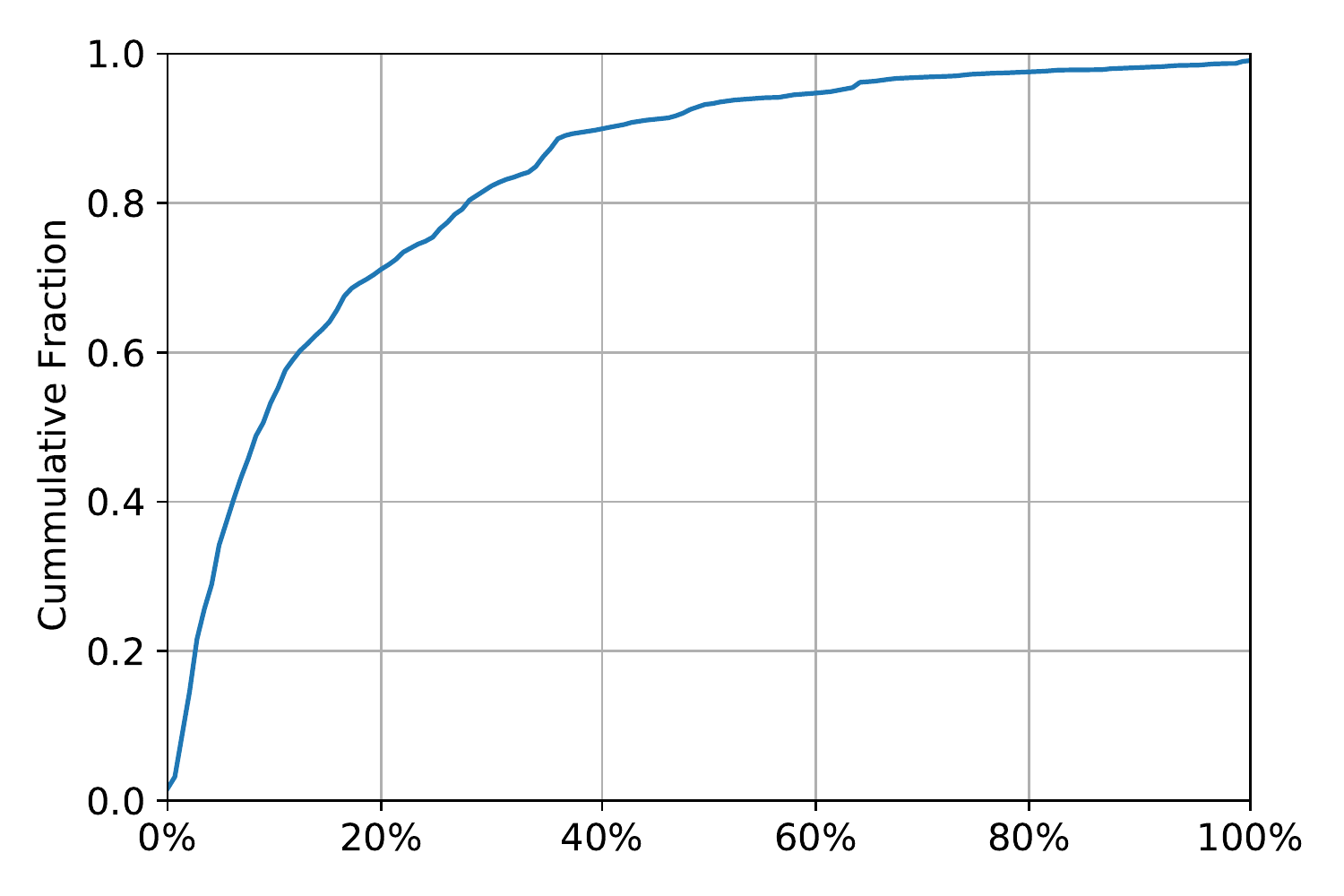}
    \caption{Memory capacity usage}
    \label{fig:intro:mem-analysis}
  \end{subfigure}
  \hfill
  \begin{subfigure}[t]{0.32\textwidth}
    \centering
    \includegraphics[width=\textwidth]{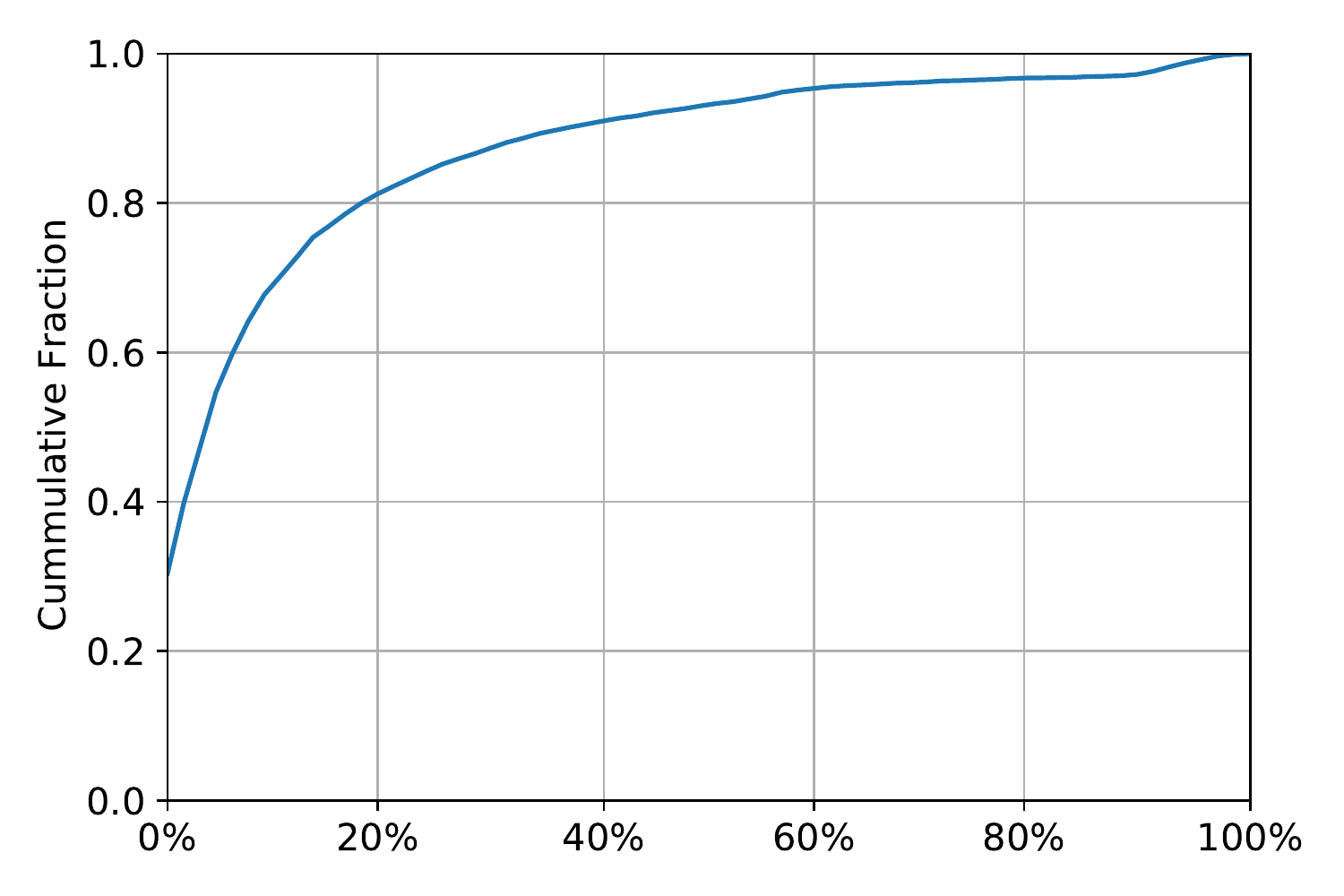}
    \caption{Memory bandwidth usage}
    \label{fig:intro:membw-analysis}
  \end{subfigure}
  \vspace{-5pt}
  \caption{CDFs of normalized ML host resource usage for over 73k colocated processing jobs running in Google datacenters over a 24 hour period. The takeaway is that host resource requirements vary widely across ML jobs.} 
  \label{fig:intro:job-analysis}
\vspace{-10pt}
\end{figure*}

The ubiquity of machine learning (ML) has led to widespread deployment of specialized hardware to accelerate ML computations (e.g., GPUs and TPUs). While vastly improving the performance per watt of ML training and inference~\cite{tpuv1, tpuv2v3-cacm}, hardware accelerators are significantly more expensive than traditional CPU servers~\cite{aws-pricing, google-cloud-pricing}. Hence, achieving energy and cost efficiency benefits requires keeping hardware accelerators highly utilized. %

Operating ML accelerators at high utilization requires feeding fresh batches of preprocessed data at the rate at which ML computations executing on accelerators request new data. Data preprocessing typically executes on CPU host resources of ML servers, as data transformations consist of user-defined functions~\cite{murray2021tfdata}. This makes it right-sizing host CPU/RAM (for data preprocessing) to ML accelerator cores (for ML model computations) imperative, in order to avoid data preprocessing stalls and maximize resource efficiency. 

However, ML jobs require diverse ratios of host CPU/RAM and ML accelerator resources~\cite{graur2022cachew, coordl}. Figure~\ref{fig:intro:job-analysis} shows the  distributions of host CPU and memory resource requirements for data preprocessing across ML jobs at Google, normalized to the peak usage for each resource. As the distributions are heavy-tailed, picking a particular CPU and memory configuration for ML accelerator hosts satisfy the requirements of a limited set of jobs at a particular point $p$ on the x-axes of the CDFs. This would leave many jobs on the left of $p$ underutilizing CPU/RAM and many jobs on the right of $p$ with insufficient CPU/RAM for data preprocessing, idling ML accelerators. Hence, provisioning ML jobs with a one-size-fits-all ratio of host CPU/RAM and ML accelerator resources is inefficient for most jobs. Meanwhile, datacenter providers must typically limit server heterogeneity to simplify resource management and server maintenance at scale~\cite{disagg-flash}.

\begin{figure}[]
\includegraphics[width=0.98\linewidth]{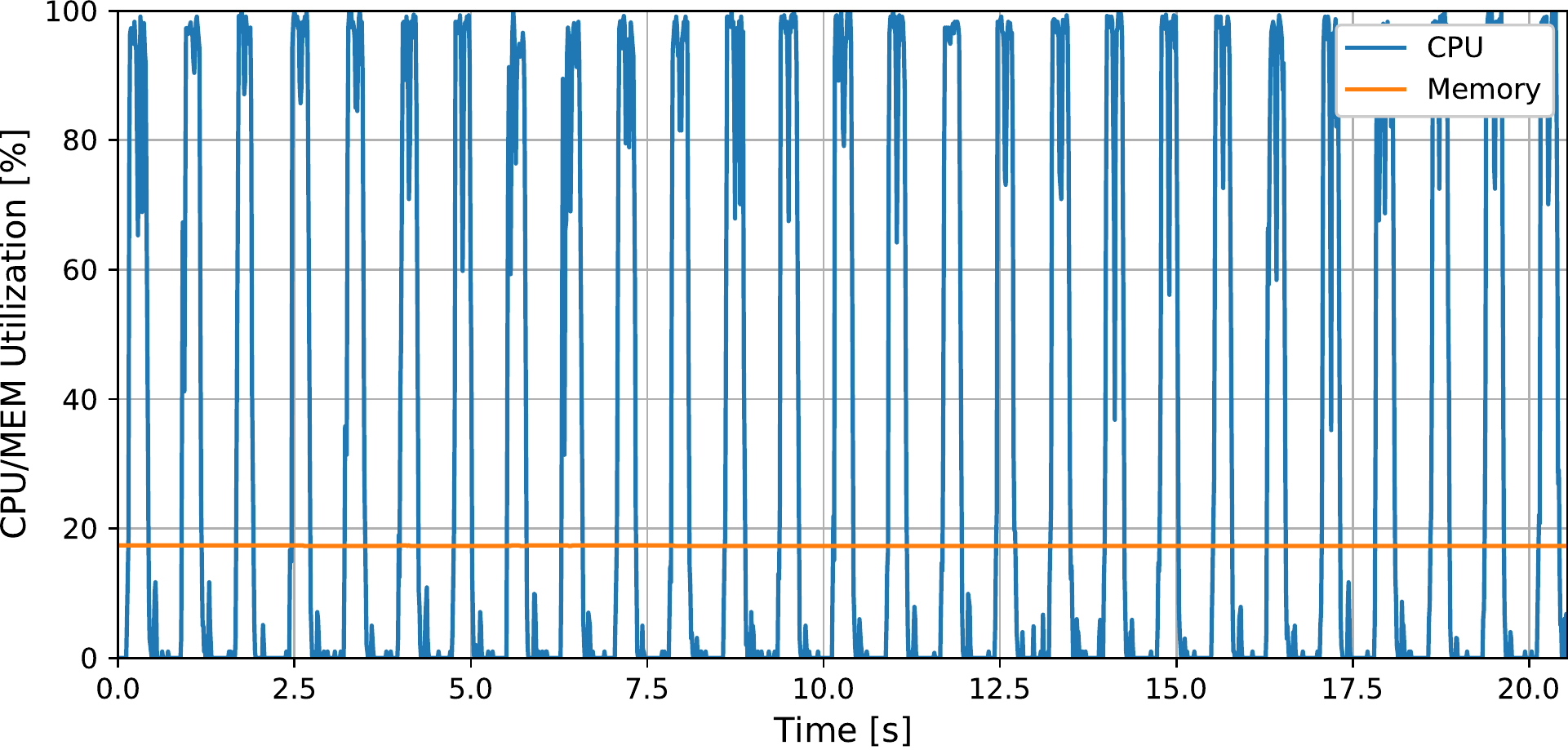}
\vspace{-5pt}
\caption{RetinaNet~\cite{lin2017focal} CPU/MEM usage when training on COCO~\cite{coco} using a GCP TPU v2-8 VM~\cite{google-cloud-tpu-config}.}
\label{fig:intro:bursty}
\vspace{-10pt}
\end{figure}

As today's ML accelerator servers are typically provisioned with generous amounts of CPU and RAM~\cite{google-cloud-tpu-config, aws-ec2-config}, one could improve resource utilization by loaning spare CPU/RAM on ML accelerator hosts to other CPU workloads. However, colocating workloads with ML data preprocessing is challenging. Figure~\ref{fig:intro:bursty} shows that data preprocessing has highly bursty CPU usage as batches of data are processed and loaded to the accelerator. Even if the average CPU utilization on many ML accelerator hosts may be low, the bursty nature of ML data preprocessing makes it difficult to share host resources with other jobs without significant interference.

An alternative approach is to \textit{disaggregate} resources for data preprocessing and ML computations. Disaggregation enables allocating CPU/RAM and ML accelerators independently for each job to meet its unique requirements. Just as disaggregating storage from compute is known to improve resource efficiency in datacenters~\cite{disagg-flash, snowflake, legoos}, disaggregating ML accelerators from CPU/RAM enables right-sizing resources for ML workloads. For example, Meta has reported that their internal closed-source ML data preprocessing system, DPP~\cite{dpp}, relies on scaling out data preprocessing to disaggregated CPU hosts to meet the data ingestion demands of large-scale recommender system model training~\cite{dpp-blog}. However, disaggregating ML data processing does not come for free: it requires recruiting potentially many remote nodes for data processing, dealing with failures of these distributed nodes, and sending large volumes of preprocessed data over the network. There is a need for a data processing platform that optimizes these tradeoffs to support resource-efficient ML input data preprocessing.

We present the design and implementation of the first open-source disaggregated data processing framework, \textit{\tfdataservice}, available through the Tensorflow GitHub project~\cite{tensorflow-github}. 
We evaluate tf.data service on a variety of production ML workloads at Google from vision and natural language processing (NLP) domains. We show that the primary advantage of disaggregation is the ability to \textit{horizontally scale-out} data processing to remote workers to remove data stalls, which improves training throughput by 31.7$\times$ and reduces cost by 26.2$\times$, on average, for input-bound jobs. %
The cost savings of maximizing accelerator utilization and reducing training time significantly outweigh the cost of using extra CPU hosts for data processing.

We also show that the disaggregated architecture of tf.data service has two additional advantages. %
First, enabling ML jobs to fetch preprocessed data from remote hosts enables sharing intermediate preprocessing results between a set of jobs that execute in overlapping time windows. This \textit{ephemeral data sharing} feature saves CPU cycles by avoiding redundant computations. %
Second, the disaggregated system architecture of tf.data service enables coordinating data ingestion across clients in distributed ML training jobs to reduce stragglers and improve training time and cost. For example, NLP models typically ingest data with a wide variety of input sizes, which can lead to stragglers among training clients. The \textit{coordinated reads} feature in tf.data service allows all ML accelerator clients in a job to fetch data from remote workers that prepare batches of data with similar size per training iteration, resulting in up to 2$\times$ overall speedup due to less data padding and synchronization overheads.

Contrary to the conventional wisdom of enforcing strict data ordering guarantees and ensuring that each example is visited exactly once per training epoch~\cite{mohan2021checkfreq, coordl, Goodfellow-et-al-2016}, we design the data sharing and coordinated reads features of tf.data service with more relaxed data visitation guarantees. We find that relaxing data visitation to at-most-once (instead of exactly-once) guarantees generally has negligible impact on model accuracy, as models are typically trained with larger datasets than in academic settings. Relaxing data visitation guarantees simplifies the design of ephemeral data sharing, coordinated reads, and fault tolerance mechanisms (e.g., compared to other ML data caching systems~\cite{graur2022cachew,coordl,quiver}).

We present the first open-source disaggregated data processing framework for ML workloads and show it has three key advantages for large-scale ML training jobs: 1) horizontal scale-out to right-size CPU/RAM host resources for data processing, 2) ephemeral data sharing to avoid redundant preprocessing among fully or partially concurrent jobs, and 3) coordinated reads to avoid stragglers in distributed training that can arise due to differences in input data sizes. Our design is inspired by lessons from deploying tf.data service in production, including the ability to relax data visitation guarantees without impacting model accuracy. Since launching tf.data service open source in 2020, the system has already been used as the foundation for several research projects, including data echoing~\cite{data-echoing}, Cachew~\cite{graur2022cachew}, and FastFlow~\cite{fastflow}. %

\section{Background and Related Work}
\label{sec:preliminaries}

\paragraph{ML input data processing} Input data processing involves an ``extract transform load'' (ETL) pipeline. Jobs read source data from storage, process data on-the-fly, and load batches of transformed data to ML accelerators for model training or inference. Common data transformations include decompressing data, parsing file formats, extracting features, and batching elements. It is also common to add randomness to input data (e.g., randomly sample, augment, and shuffle elements) to improve model generalization~\cite{autoaugment, randaugment, imagenet, best-practices-cnns}.

While model training or inference typically executes on expensive, specialized hardware, user-defined input data transformations  execute on general purpose CPUs. Their relative cost difference makes it particularly important to  ensure that accelerators operate at high utilization~\cite{aws-pricing, google-cloud-pricing}. If the ML computation does not have to wait for data, the job is considered \emph{model-bound}. Otherwise, the job is considered \emph{input-bound}. Input-bound jobs are more problematic as they leave valuable ML hardware underutilized, however model-bound jobs can also be costly by leaving CPUs underutilized. Removing input bottlenecks is critical as input-bound jobs hog valuable hardware for extended periods of time, incurring significant costs and significant delays for the job itself as well as other jobs that are waiting for the hardware resources to free up. %

\paragraph{Data processing frameworks} \review{2}{Generic data processing frameworks, such as Beam~\cite{beam}, Flume~\cite{flume} and Spark~\cite{spark} are often used for \textit{offline} ML data processing (i.e. processing that takes place prior to any ML compute and handles data cleaning, feature engineering, normalization, etc.). These generic frameworks are not suitable for \textit{online} ML data processing (i.e. on-the-fly data processing during a training job that handles data augmentation, shuffling, batching, etc.), primarily due to the overhead they impose. For instance, Spark Streaming recommends batching work at 50ms granularity~\cite{spark-streaming-performance} while ML training jobs often have step times below 1ms. Other issues stem from API mismatches between generic data processing frameworks and ML training frameworks, a lack of holistic optimization opportunities to improve runtime performance (as the preprocessing and ML compute frameworks are disjoint), and the difficulty of accommodating ML-specific preprocessing requirements (such as relaxed data visitation guarantees, which we describe below). %
}

ML-specific data frameworks, used for online data preprocessing, include PyTorch DataLoader~\cite{pytorch-dataloader}, NVIDIA DALI~\cite{nvidia-dali}, and tf.data~\cite{murray2021tfdata}. We focus on tf.data as it is widely used internally at Google and in open-source Tensorflow programs. tf.data provides an API and a runtime for creating and executing efficient ML data processing pipelines. With tf.data, users can express and execute input data pipelines using a variety of libraries that integrate with ML frameworks like PyTorch~\cite{pytorch} and Tensorflow~\cite{tensorflow}. It provides generic operators (e.g. \inline{map}, \inline{filter}, \inline{prefetch}, etc.) and an autotuning feature that automatically adjusts runtime configuration knobs (e.g. parallelism, prefetching, etc.)~\cite{tfdataperformance}.

\paragraph{Colocated vs. disaggregated data processing} 
Traditionally, ML frameworks have \textit{colocated} input data processing and ML computations on the same machine~\cite{pytorch-dataloader, mxnet_dataio, tfdata}. However, feeding powerful modern accelerators with data at sufficient throughput requires  more CPU and memory resources than are often locally available~\cite{dpp, graur2022cachew}. This motivates a \textit{disaggregated} processing mode, where data processing executes on separate machines whose resources can be scaled independently from expensive ML accelerators. The ability to right-size resource allocations is critical to satisfy the wide variety of resource requirements in ML jobs (see Figure~\ref{fig:intro:bursty}) and ensure that all allocated resources --- both costly ML accelerators and CPU/MEM --- remain highly utilized.

\paragraph{Data visitation guarantees} In the ML community, it is customary to train models with \textit{exactly-once} data visitation semantics for each epoch~\cite{Goodfellow-et-al-2016,bottou2010large, bottou-curiously-fast-convergence,mohan2021checkfreq}. This means that every sample in the dataset is visited exactly once before any sample is re-visited during training. %
For small datasets, deviating from exactly-once guarantees can skew the data distribution, potentially leading to a less generalizable or lower accuracy model~\cite{mohan2021checkfreq,Goodfellow-et-al-2016}. %
In \S\ref{sec:design}, we will discuss how relaxing data visitation guarantees is possible for production ML jobs, as they generally train on significantly larger datasets that are continuously updated~\cite{dmlsbook2022}.%

\paragraph{Prior work on optimizing input processing} 
Several approaches have been proposed to alleviate input data stalls in ML training jobs. Plumber~\cite{plumber} and tf.data's autotuning harness~\cite{tfdata} dynamically tune software parallelism and memory buffer sizes to maximize performance on a given training node. However, this tuning does not scale resources beyond a single node. NVIDIA DALI supports offloading data processing to GPUs~\cite{nvidia-dali}. This alleviates CPU bottlenecks but may lead to GPU resource contention among input data transformation tasks and model training tasks. Other works cache and reuse (transformed) data across epochs~\cite{data-echoing, revamper} or across jobs~\cite{graur2022cachew, quiver, coordl}, trading storage capacity to save CPU cycles for repeated data processing. However, caching solutions are still not guaranteed to eliminate data stalls. In-memory caching solutions have capacity limitations since ML datasets often exceed the size of a training node's RAM~\cite{tfdata}. SSD-based caching requires high parallelism to avoid I/O bottlenecks when reading large volumes of data from storage.

Meta's closed source Distributed Data Processing (DPP) service~\cite{dpp} proposes horizontally scaling data workers with disaggregated data processing. Zhao et. al~\cite{dpp} characterize the compute, memory, and network resource requirements of data processing compared to model training for recommender system models at Meta and advocated for disaggregation to scale-out data processing and avoid input bottlenecks. However, they do not quantify the performance and cost benefits of disaggregated data processing compared to colocated data processing. In contrast, we provide the first quantitative analysis of how much disaggregated data processing actually improves the overall performance and cost of production ML workloads. We also go beyond DPP's design and show the benefits of disaggregation besides horizontal scaling, such as enabling ephemeral data sharing and coordinated reads. %

FastFlow~\cite{fastflow} builds on top of tf.data service, leveraging its ability to preprocess data on both local and remote resources. FastFlow extends our disaggregation mechanism to support splitting data preprocessing between local and remote workers at specific points in an input pipeline. The system selects a pipeline split that maximizes throughput in a fixed-size tf.data service deployment, i.e., the system decides which portion of an input pipeline to process locally vs. remotely. %
FastFlow is complementary to our work and further shows the benefits that flexible resource allocation with disaggregation can bring to ML data processing.

\section{\tfdataservice Design}
\label{sec:design}

We present \tfdataservice, a system that enables disaggregating and distributing data processing for ML workloads. tf.data service integrates seamlessly with ML frameworks, such as TensorFlow or PyTorch, and is particularly designed to meet the data processing needs of ML training jobs. %
We design \tfdataservice with three key principles in mind:

\paragraph{(1) Eliminate input data stalls} Eliminating data stalls is key to maximizing throughput, which is an important performance metric for ML training jobs. %
As we will demonstrate in \S\ref{sec:eval}, using extra CPU/RAM resources to alleviate data preprocessing stalls is often cost-efficient for input-bound jobs as it helps them complete faster and hence consume expensive hardware accelerators for less time.

\paragraph{(2) Avoid redundant data preprocessing} Input data pipelines are often repeatedly executed in ML clusters, for example in hyperparameter tuning or model search workflows~\cite{murray2021tfdata, oneaccess}. Since data preprocessing can be compute, memory, and power-intensive~\cite{murray2021tfdata, dpp}, reducing redundant data preprocessing (i.e., sharing intermediate results) across jobs is important for hardware and energy efficiency. %

\paragraph{(3) Simplify the design by relaxing constraints}
While guaranteeing exactly-once data visitation is customary when benchmarking ML training on academic datasets~\cite{mohan2021checkfreq, coordl}, we find that production ML training jobs allow for more relaxed data visitation guarantees with negligible impact on accuracy.
These jobs train on vast volumes of data that are continuously updated~\cite{dmlsbook2022,Goodfellow-et-al-2016}. Some datasets are large enough that training on all data may not be practical as the model can safely converge on a subset of the entire dataset without the risk of overfitting~\cite{Goodfellow-et-al-2016}. %
Hence, we find that \textit{at-most-once} data visitation is sufficient for most jobs (i.e., each sample is visited at most once per epoch) \footnote{Further relaxing visitation guarantees to \textit{zero-once-or-more} can produce a high quality model in some cases, as long as the random transformations applied to the source data to ensure a sufficiently diverse and representative distribution~\cite{data-echoing,revamper}.}.
We can leverage relaxed data ordering guarantees to simplify fault-tolerance.%

\smallskip

With the above principles in mind, we design tf.data service as a disaggregated system that enables: eliminating input data bottlenecks by horizontally scaling out data workers (\S\ref{sec:design:arch}), reducing CPU usage for data processing by sharing ephemeral data transformations between jobs  (\S\ref{sec:data-service-sharing}), and avoiding stragglers in distributed training by coordinating reads between multiple clients (\S\ref{sec:coordinated-reads}).

\subsection{System Architecture}
\label{sec:design:arch}
\begin{figure}[]
\includegraphics[width=\linewidth]
{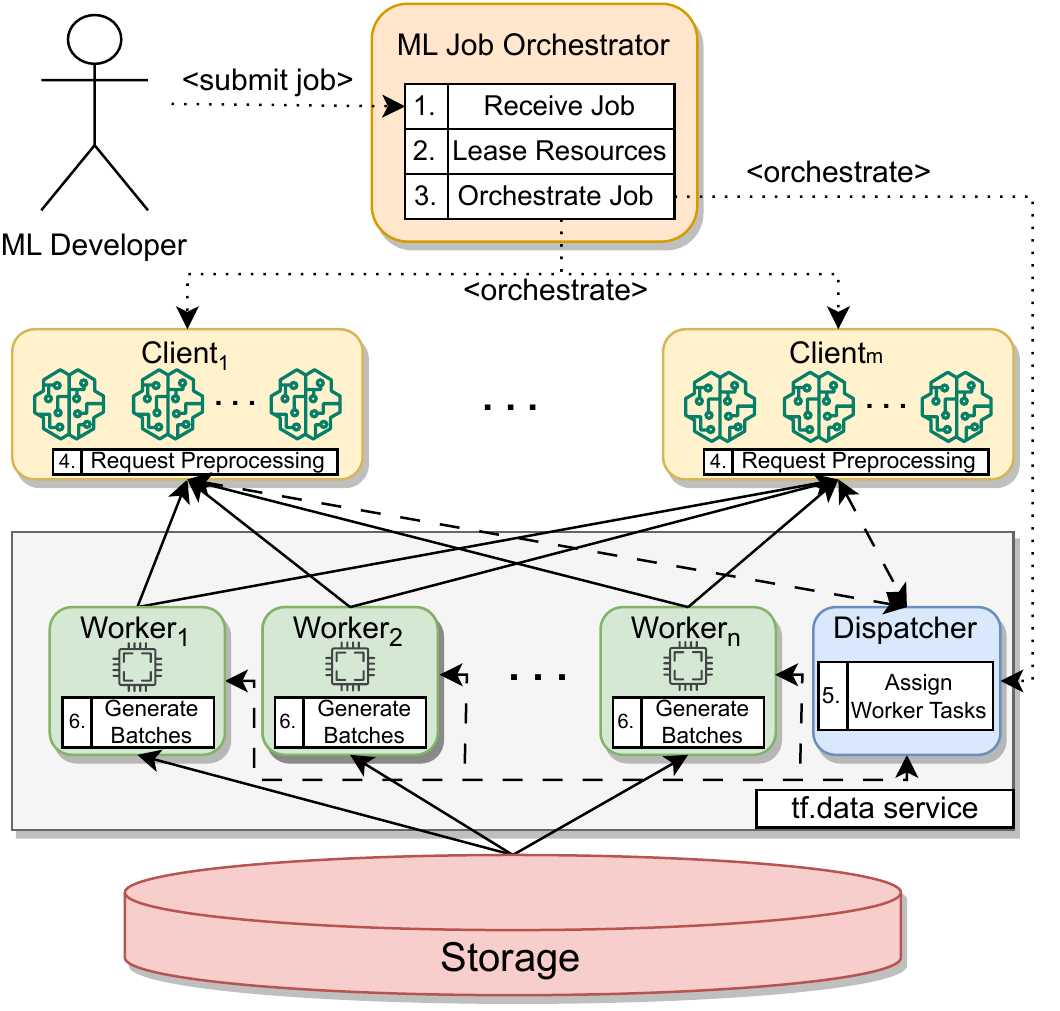}
\caption{Architecture and workflow. Solid lines represent the data path, dashed lines represent the control path, and dotted lines the execution flow.}
\label{fig:tfdata-service-architecture}
\vspace{-10pt}
\end{figure}

Figure~\ref{fig:tfdata-service-architecture} shows the \tfdataservice architecture. The core system is composed of a \emph{dispatcher}, which manages various metadata, and a pool of \emph{workers}, which perform data preprocessing. The dispatcher and workers are managed by an ML job \emph{orchestrator}. \textit{Clients} execute ML computations (e.g., model training) on servers equipped with accelerators. Distributed training jobs consist of multiple clients. Source data is stored in a distributed storage system (e.g. Colossus~\cite{colossus}) or object storage (e.g. GCS~\cite{google-cloud-storage}) accessible by all workers. %

\paragraph{Life of a job} As shown in Figure~\ref{fig:tfdata-service-architecture}, 
the ML developer first submits their ML job to the orchestrator. %
The orchestrator spins up the tf.data service dispatcher and workers. Clients then register their input data pipeline with the dispatcher, using the API we describe in \S\ref{sec:api}. The dispatcher distributes a \tfdata computation graph representing the input data pipeline to all available workers. This informs workers which source data to read from the storage layer and how to preprocess data into batches. For each job, each worker processes the same input data pipeline but on different partitions of the source dataset, determined by the dispatcher's sharding policy. The dispatcher informs clients about the IP addresses of workers and clients request batches from workers over the network.  All communication between the clients, the dispatcher, and the workers is done via gRPC, which uses HTTP/2, and multiplexes multiple calls on a single TCP connection~\cite{grpc-documentation}. %

\paragraph{Dispatcher} The dispatcher  manages metadata about registered workers, clients, and input pipelines. It receives heartbeats from clients and workers, assigns dataset processing tasks to workers for partitions of the input dataset using a sharding policy (see \S\ref{sec:design:sharding}), and notifies clients of worker pool updates. It also keeps track of states of dataset processing tasks and notifies clients of completed tasks. To avoid bottlenecks, the dispatcher does not perform any data processing.

\paragraph{Workers} Workers execute data processing tasks defined in a \tfdata computation graph. Workers read data from storage, apply requested transformations, and store the samples in a buffer. Workers respond to client requests for buffered samples. %
The number of workers and their CPU/RAM resources can be specified manually or tuned automatically by integrating \tfdataservice with an auto-scaling framework~\cite{graur2022cachew,rzadca2020autopilot}.

Workers are logically disaggregated components, however they can be deployed on remote CPU servers or local host resources on clients. 
Remote workers serialize data before sending it to clients over the network, hence clients need to deserialize the data before copying it to accelerator memory. In bandwidth constrained environments, workers can also compress the data before serialization and clients can decompress the data. In deployments where bandwidth is abundant, we find that compression is not helpful as it occupies unnecessary CPU cycles for both clients and workers. In contrast to remote workers, local workers execute as part of the client's main TensorFlow process and hence can communicate directly via function calls with the client. Remote workers are useful for scaling out data preprocessing to eliminate input bottlenecks when local host resources on client machines are not sufficient. Local workers are useful for jobs with relatively low data ingestion rate requirements that run on client machines with abundant CPU and memory. For jobs that rely on many remote workers to eliminate input data bottlenecks, local CPU and memory resources on clients are consumed by deserialization, decompression, and data loading to accelerators, leaving few spare resources for preprocessing data with local workers.

\paragraph{Clients} Without \tfdataservice, each client would run its own local input pipeline for data preprocessing. With \tfdataservice, clients instead send gRPC requests to workers to fetch preprocessed data. To maximize data ingestion, clients can request data from multiple workers in parallel. Each client stores results in a client-side buffer and returns batches to the downstream ML computation when requested. %

\paragraph{Orchestrator} At Google, we use Borg~\cite{verma2015borg,tirmazi2020borg} as the \tfdata service orchestrator. 
Borg deploys clients, workers, and the dispatcher as containers on machines in the same Borg cell. In our fleet, client machines with accelerators are dedicated to the ML job whereas the \tfdataservice workers run on multi-tenant machines with fungible resources.  %
Borg uses Autopilot~\cite{rzadca2020autopilot} to horizontally scale the number of worker nodes based on user hints and CPU utilization. %
Autopilot can also vertically scale the resources allocated to each worker or dispatcher container using machine learning and other statistical techniques that leverage raw hardware utilization signals~\cite{rzadca2020autopilot}. %
tf.data service can also be deployed with other orchestration systems, such as Kubernetes~\cite{kubernetes}. %
Kubernetes supports horizontal scaling~\cite{kubernetes-hpa} vertical scaling~\cite{kubernetes-vpa}. Users and researchers can also implement custom autoscaling policies for \tfdataservice. For example, Cachew~\cite{graur2022cachew} autoscales workers based on batch time processing monitoring.

\subsection{API and Pipeline Optimizations}
\label{sec:api}

\begin{figure}[tb]
\begin{lstlisting}[style=Python]
ds = make_dataset()
ds = ds.distribute(
    processing_mode=ShardingPolicy.OFF,
    service_address=...)
for <@batch@> in ds:
    train_step(<@batch@>)
\end{lstlisting}
\caption{\tfdataservice API example.}
\label{fig:tf-data-service-api}
\vspace{-13pt}
\end{figure}

\paragraph{API} 
Figure~\ref{fig:tf-data-service-api} shows an example of a simple \tfdataservice pipeline. Here, \inline{make\_dataset()} (line 1) is a user-defined function that generates \tfdata dataset object using the standard tf.data API. To indicate that a \tfdata dataset should use \tfdataservice, we apply the \inline{distribute} transformation to the dataset definition (lines 2-4). This serializes the upstream dataset definition and sends it to the dispatcher. The dispatcher shares the definition with the pool of workers and sends the worker IP addresses to the client. During input pipeline execution -- using the \inline{for} loop (lines 5-6) -- the client sends RPCs to the workers to request data. This simple Python API abstracts communication between clients and workers. User programs simply iterate over the Python dataset object the same way they would for colocated processing with \tfdata.

Via the \inline{distribute} transformation users specify several deployment parameters, such as the compression method to use (if any) when sending data over the network, whether clients should fetch batches from local and/or remote workers (by default, they read from both), and the communication protocol between clients and workers (by default gRPC). %

\paragraph{Pipeline Optimizations} 
tf.data service pipelines benefit from all optimizations available in tf.data, such as static graph optimization passes and dynamic autotuning~\cite{murray2021tfdata}. Before the client registers an input pipeline with the dispatcher, it undergoes a number of optimization stages that, among others, try to eliminate dead transformations, inject transparent prefetching, and fuse operators (e.g. \inline{map}-\inline{filter} fusion) to reduce software overhead. Besides the optimizations that act on the input pipeline's dataflow graph prior to execution, input pipelines also transparently benefit from automatic tuning of operator-specific parameters at runtime (e.g., the number of software threads to use), through the \texttt{AUTOTUNE} mechanism~\cite{murray2021tfdata}. These optimizations transparently tune the level of parallelism for each operator, as well as other parameters (e.g. size of a prefetch buffer) that explicitly use the \texttt{tf.data.AUTOTUNE} token. These optimizations take place at runtime since they are dependent on the target hardware configuration and the underlying data distribution.

\subsection{Source Data Sharding} 
\label{sec:design:sharding}

When using tf.data service, users can specify a sharding policy for the \inline{distribute} transformation (line 3 in Figure~\ref{fig:tf-data-service-api}). Sharding policies dictate how source data should be partitioned across workers. Source data is typically stored in several files, hence, each file constitutes a source data shard. It is also possible to change the granularity of sharding to individual samples in the source data or to sets of files~\cite{tfdataserviceapidocumentation}. tf.data service offers several sharding policies, including: (1) no sharding via the \inline{OFF} policy (2) disjoint first-come-first-served sharding via the \inline{DYNAMIC} policy, and (3) several static sharding policies that pre-allocate shards across workers.

With the \inline{OFF} sharding policy %
the dispatcher does not shard the dataset and 
each worker processes the entire dataset independently. Each worker typically processes data in a different random order. This sharding policy is suitable for models that are robust to the most relaxed visitation guarantees, \textit{zero-once-or-more}. The benefit of this approach is that it does not require dataset partitioning and coordination of partition assignments across workers.  

The \inline{DYNAMIC} sharding policy targets jobs which require some data visitation guarantees. With dynamic sharding, the dispatcher partitions the source dataset into disjoint shards. The workers request shards from the dispatcher whenever they run out of data to process. Workers process one shard at a time. Since the shards are completely disjoint, each source data element contributes to at most one batch which is seen by at most one client. By using a larger number of shards than the number of workers, the dispatcher can load balance across the workers. If workers are preempted or fail, dynamic sharding provides \emph{at-most-once} visitation guarantees, as distributed shards are not recovered until the following epoch (see \S\ref{sec:design:fault-tolerance}). In the absence of preemptions and failures, dynamic sharding provides \emph{exactly-once} visitation guarantees. %

tf.data service also supports static sharding strategies, which assign source data partitions to the available workers upfront at the beginning of a job. Users are also free to implement their own custom sharding policies.%

\subsection{Fault tolerance} 
\label{sec:design:fault-tolerance}

Disaggregation allows \tfdataservice to horizontally scale data processing across multiple machines, which introduces a new failure domain. The likelihood of a data worker failing increases as we distribute data processing to more workers. We use a simple recovery mechanism for workers, enabled by  relaxed visitation guarantees tolerated by production ML jobs. We also discuss dispatcher fault tolerance. 

\paragraph{Worker fault tolerance} To simplify fault tolerance, \tfdataservice workers are designed to be stateless. 
Hence, a restarted worker will register with the dispatcher and retrieve the dataset definition it needs to process, like any new worker would. 
Worker failure recovery depends on the sharding policy. With no sharding, restarted workers re-process the entire dataset in a newly randomized order. 
 In the dynamic sharding case, the restarted worker queries the dispatcher for the next split to process. The dispatcher assigns each split to exactly one worker. If the worker fails when it is processing a split, the remaining data in that split is lost. Hence with dynamic sharding, \tfdataservice guarantees samples are seen at-most-once. We opted for this design due to its simplicity and performance. %
Exactly-once visitation guarantees can be implemented if desired. This would require the dispatcher to log the distribution of shards and the workers to persist how much of the shard has been processed (at operator granularity). The same or a new worker can then be restarted with the same shards and the computation can be resumed with the same state of the original worker. 

\paragraph{Dispatcher fault tolerance} The dispatcher writes its state changes to a write-ahead journal, including registered datasets, active workers, and active clients. On restart, the dispatcher replays the journal to restore its previous state. When down, the dispatcher cannot register new jobs, and, in case any sharding is used, cannot distribute new shards to workers. Workers will continue to produce batches for their active jobs until their data shards are fully consumed, or if no sharding is used, workers will process the entire dataset. Clients will continue to train on any incoming batches. If the dispatcher downtime is short, active jobs will remain unaffected.

\subsection{Ephemeral Data Sharing}
\label{sec:data-service-sharing}

\begin{figure}[]
\centering
\includegraphics[width=\linewidth]{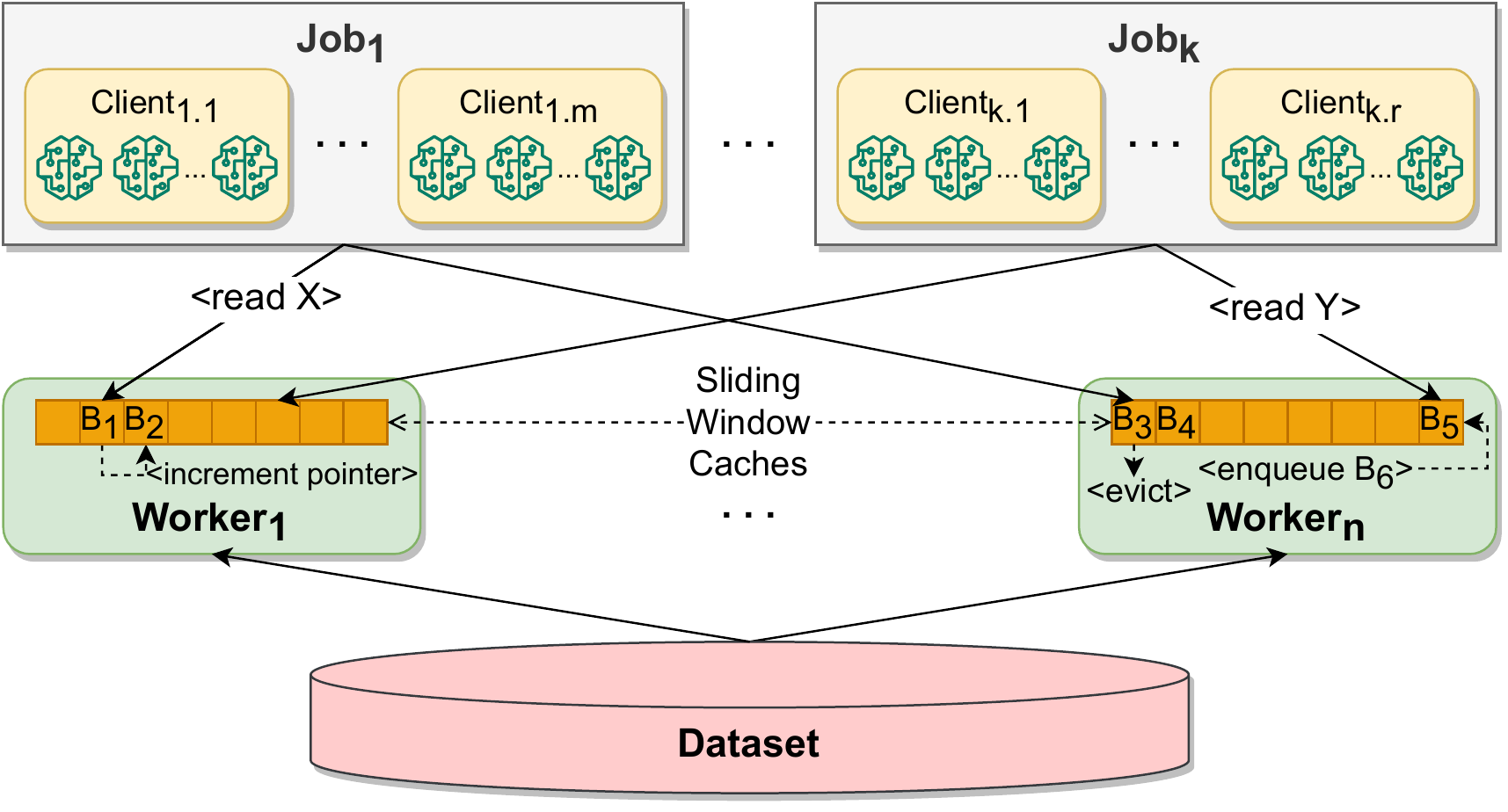}
\caption{Ephemeral data sharing workers serve requests from different jobs via sliding window caches.}
\label{fig:shared-tf-data-service}
\vspace{-10pt}
\end{figure}

Disaggregating data preprocessing enables sharing intermediate preprocessing results between multiple training clients. Sharing is useful as input data pipelines are frequently re-executed across jobs, for example in hyperparameter tuning or model search training workflows~\cite{oneaccess}. For example, a previous study at  Google found that 10\% of unique input data pipelines executed over a one month period in the fleet accounted for 77\% of input pipeline executions~\cite{murray2021tfdata}. %

To capitalize on the opportunity of reusing preprocessed data across jobs that execute the same input data pipeline, we implement the \emph{ephemeral data sharing} functionality that allows concurrently executing jobs to use a shared \tfdataservice deployment instead of reading and preprocessing the same data in separate deployments. Figure~\ref{fig:shared-tf-data-service} illustrates the data sharing process. Each worker stores a sliding window cache of batches it produces. The worker additionally stores a pointer in this cache for each job that it supplies data to. When a job requires a data batch from a worker, the worker returns the batch in the cache at the location pointed to by the job's pointer, then increments the pointer. For example, in Figure~\ref{fig:shared-tf-data-service}, when \texttt{Job$_1$}'s client issues \texttt{read X}, it fetches batch \texttt{B$_1$} from \texttt{Worker$_1$}'s sliding window cache. After the read, \texttt{Worker$_1$} increments \texttt{Job$_1$}'s pointer, moving it to batch \texttt{B$_2$}.

Jobs whose pointers are at the front of the cache dictate the production of new data and eviction of old data. In this case, the cache functions like a queue. When such jobs ask for a batch, they implicitly cause the worker to compute a new batch that is added to the front of the cache after the read is complete. To limit the RAM used for the cache, the batch at the back of the cache is evicted. Slower jobs pointing to the back of the cache will not be able to see the discarded batch. After the eviction, their pointers remain unchanged and implicitly point to the end of the queue, while all other pointers are decremented in order to continue pointing to the correct batches. For instance, in Figure~\ref{fig:shared-tf-data-service}, when \texttt{Job$_k$}'s \texttt{read Y} operations fetches batch \texttt{B$_5$}, this causes \texttt{Worker$_n$} to compute and enqueue a new batch \texttt{B$_6$}, and push every other batch back. Due to the limited cache space, \texttt{B$_3$} is evicted and \texttt{B$_4$} is pushed to the end of \texttt{Worker$_n$}'s sliding window cache. \texttt{Job$_1$} will not get to train on \texttt{B$_3$}, as it is permanently evicted, and will implicitly point to \texttt{B$_4$}.

\review{3}{In extreme cases, one job can severely lag behind another, thus facing high eviction rates. This could degrade accuracy for models trained on very small datasets, due to overfitting. In practice, we have not seen this issue as datasets are rarely that small; many production datasets are even too large to fully visit during training, thus comfortably accommodating evictions. Moreover, random augmentations and shuffling, which are frequently present in input pipelines, artificially increase the diversity and size of the dataset, making such scenarios even more rare~\cite{geiping2023data,revamper}.}

To analyze the benefits of data sharing, let us assume that the data processing cost for each job in a hyperparameter tuning experiment using the same input data pipeline is $C$. If all jobs run at the same speed, then the data processing cost with sharing is $C$. Without sharing, the data processing cost would be $k \times C$. If the jobs run at different speeds, slower jobs may miss some preprocessed batches due to cache evictions needed to accommodate faster jobs. In the worst case, the jobs run sequentially and each job only shares the final window populated by the previous job (the preprocessing for the rest of the batches needs to be re-executed since it is no longer available in the buffer). In this case, the processing cost with sharing is $k \times C - (k - 1) \times \frac{\textrm{cache size}}{\textrm{dataset size}} \times C$ (assuming constant processing cost per element).

Ephemeral data sharing differs from related work on ML data caching in several ways. First, as we relax data visitation guarantees, a large number of concurrently running jobs can share a single service deployment without stalling for slower jobs. In contrast, with exactly-once visitation guarantees, the fastest jobs would need to wait for the slowest jobs.  Second, our data sharing algorithm supports partially-concurrent, asynchronous jobs with different ML models attached (i.e., ephemeral data sharing supports jobs whose execution overlaps only partially and no synchronization is required between jobs). This contrasts with the caching approach proposed in CoorDL~\cite{coordl}, which is designed for %
fully-concurrent, synchronous, hyperparameter tuning jobs with exactly-once visitation guarantees. Third, unlike Cachew~\cite{graur2022cachew}, which caches source datasets and preprocessed datasets on SSD storage, \tfdataservice caches data in memory with a shorter time to live. This improves cache performance, but requires that job execution overlaps at least partially. %

\subsection{Coordinated Reads}
\label{sec:coordinated-reads}

\begin{figure}[tb]
  \centering
  \begin{subfigure}[t]{0.23\textwidth}
    \centering
    \includegraphics[width=\textwidth]{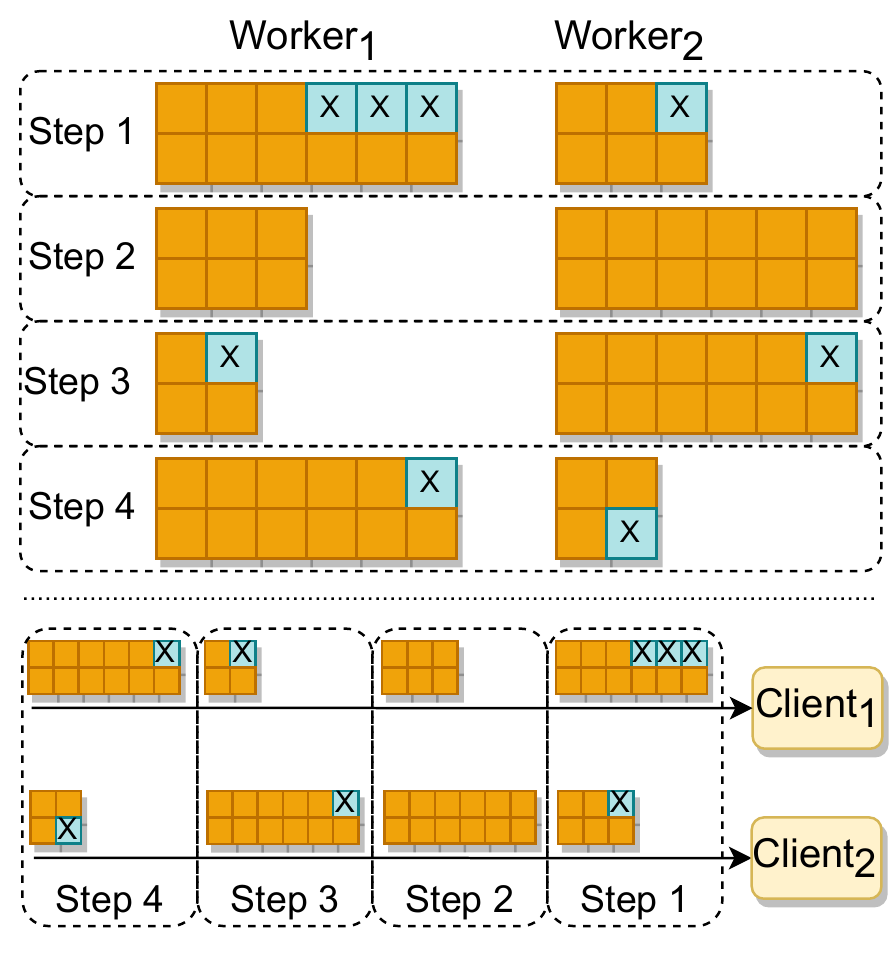}
    \caption{No coordinated reads}
    \label{fig:design:coordinated-reads-example:uncoordinated}
  \end{subfigure}
  \begin{subfigure}[t]{0.23\textwidth}
    \centering
    \includegraphics[width=\textwidth]{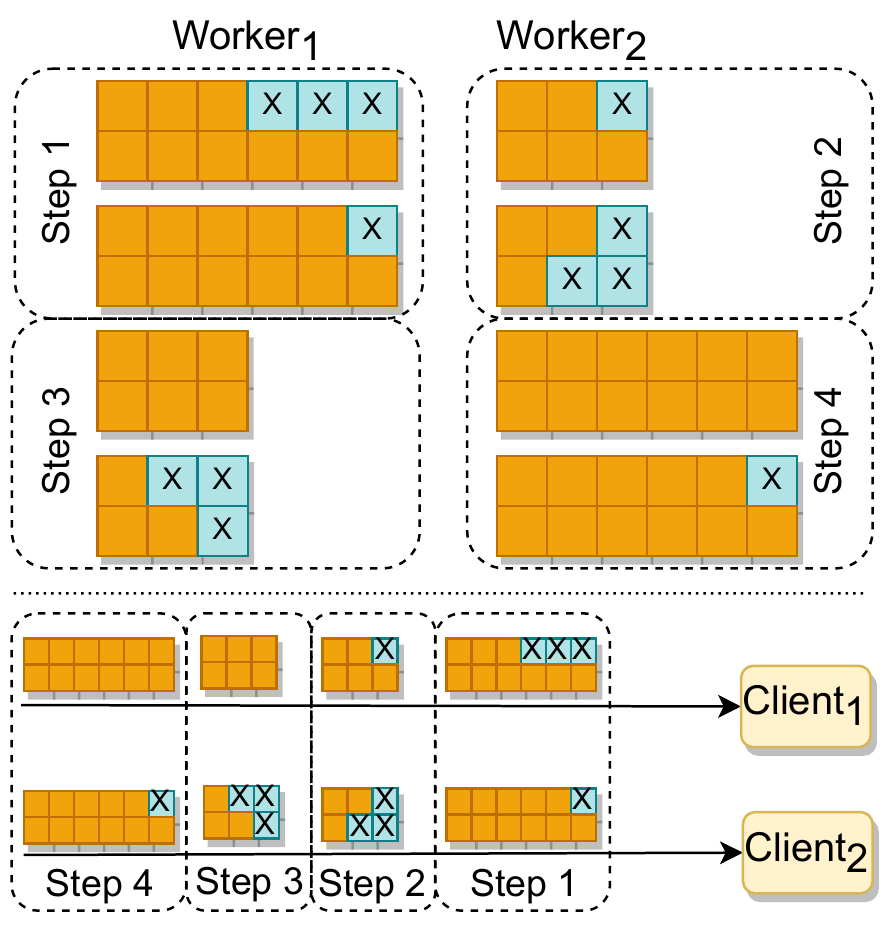}
    \caption{With coordinated reads}
    \label{fig:design:coordinated-reads-example:coordinated}
  \end{subfigure}
  \vspace{-5pt}
  \caption{
  Batch distribution to two clients with and without coordinated reads. The deployment has two workers generating batches. In (a) the batches used in a step are generated by two workers every time while in (b) workers take turns to supply batches for each step. Each batch consists of two samples. Each {\protect\includegraphics[trim = 0.1cm 0.1cm 0.1cm 0.1cm, clip=true, scale=0.5]{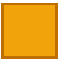}} represents a scalar value in a sample. Batch samples are padded ({\protect\includegraphics[trim = 0.1cm 0.1cm 0.1cm 0.1cm, clip=true, scale=0.5]{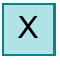}} blocks) to have a length equal to the maximum length sample in the batch. In (b), samples are grouped into buckets based on their unpadded length. In each step, the designated worker sends batches from the same bucket to the clients.} 
  \label{fig:design:coordinated-reads-example}
\end{figure}

Distributing ML model training across multiple clients is increasingly common as model and dataset sizes continue to scale~\cite{param-server, distrib-ML}. Clients in a distributed ML training job usually synchronize model parameter updates, to ensure convergence~\cite{BytePS}. Synchronization between training clients introduces another type of potential bottleneck that can leave expensive ML accelerators idle: waiting for stragglers. While stragglers can occur for a variety of reasons in distributed ML jobs, here we focus on stragglers that occur due to uneven input data sizes across clients. We show how we can leverage disaggregated data preprocessing in \tfdataservice to coordinate reads in a way that evens out input data sizes across clients to avoid stragglers in distributed ML training.

Some ML models (particularly in NLP), train on variable-sized source data and produce variable-sized preprocessed samples. A common approach is to pad the variable-sized samples to a fixed size, thus producing homogeneously shaped batches that fit the expected input shapes of the ML model. %
tf.data and TensorFlow offer a more efficient approach than padding: (1) tf.data can produce variable sized batches by padding to the longest sample in each batch (as opposed to a batch-agnostic static size) and (2) TensorFlow can dynamically kernelize its ML operations for the variable input tensor shapes it sees during training. This feature improves training time by not relying on fixed-size padding, however it is susceptible to synchronization overheads in distributed synchronous training when, within the same training step, different clients receive batches of different sizes and hence take different times to train on the batches.

In such settings, synchronous training requires careful coordination of input data batches to avoid stragglers. Our goal is to ensure that batches in each training step are roughly the same size and thus incur similar training times. 
Consider the example of training an NLP model on two clients (\texttt{Client$_1$} and \texttt{Client$_2$})  in Figure~\ref{fig:design:coordinated-reads-example} with two preprocessing workers, (\texttt{Worker$_1$} and \texttt{Worker$_2$}). As there are two clients, each training step requires two batches. %
Without coordinating reads, each training step, clients receive differently sized batches in the same training step. %
In contrast, with coordinated reads, each worker is tasked with generating $m$ similarly sized batches (where $m$ is the number of clients; $m=2$ in Figure~\ref{fig:design:coordinated-reads-example:coordinated}). In the example, we configure coordinated reads to group samples of length $(0,3]$ into one bucket, and of length $(3,6]$ into another. Workers generate batches for each bucket, padding samples to the length of the longest sample in each batch. At each training step, only one of the workers supplies batches to all clients, from the same bucket. Workers take turns in round-robin fashion to supply batches for a training step. Hence, each worker supplies $m$ batches every $n$ steps (where $n$ is the number of workers). This scales well as each worker has $n-1$ steps to prepare $m$ batches. To avoid encountering network bottlenecks, workers can send batches to clients ahead of time in predetermined round-robin client-side buffer slots. Overall, the coordinated reads feature ensures that clients get similarly sized batches with minimal padding, thus spending similar amounts of time on ML computation. As we show in \S\ref{eval:coordinated-reads}, this reduces synchronization overheads and improves training time and cost.

\begin{figure}[t]
\begin{lstlisting}[style=Python]
ds = ds.bucket_by_sequence_length(
    bucket_boundaries=[128, 256])
ds = ds.group_by_window(window_size=<@num\_consumers@>)
ds = ds.flat_map(lambda x: x)
ds = ds.distribute(
    job_name="coordinated_reads_job",
    num_consumers=<@num\_consumers@>,
    consumer_index=i, ...)
\end{lstlisting}
\caption{Dynamic-sequence-length training with \tfdataservice and coordinated read.}
\vspace{-10pt}
\label{fig:dynamic-sequence-length-training}
\end{figure}

Disaggregation is essential for enabling coordinated reads as it allows clients to communicate with and selectively read from a different worker each training step. Disaggregation also enables the service to scale out when workers are not able to keep up with the data ingestion rate of clients. %
To ensure high scalability to a large number of workers and maintain simple fault tolerance mechanisms, we design the service such that coordination only takes place within workers, not across them. This avoids additional dependencies and communication between workers.

 Figure~\ref{fig:dynamic-sequence-length-training} shows how users can configure coordinated reads in the tf.data service API. Users define the desired bucket boundaries and call \inline{bucket\_by\_sequence\_length} (lines 1-2) to bucketize the training data by sequence length. Here the bucket boundaries are $(0,128]$, $(128,256]$, and $(256,\infty)$. The \inline{group\_by\_window} call (line 3) will group batches from the same bucket into windows of \inline{num\_consumers} length. 
Clients will receive data from the same window, hence with batches of similar sequence length. The tf.data service documentation provides more details on the API~\cite{tfdataserviceapidocumentation}.
\section{Evaluation}
\label{sec:eval}

In this section we quantify the performance and cost benefits that \tfdataservice achieves for input-bound ML jobs by horizontally scaling out workers to eliminate data stalls. We also evaluate ephemeral data sharing and coordinated reads, which bring benefits even for model-bound ML jobs.

\subsection{Methodology}

\paragraph{Workloads} We evaluate \tfdataservice service on eight production models, which we refer to as $M_1, M_2, ..., M_8$. Models $M_1$ to $M_4$ are heavy users of \tfdataservice internally from the computer vision domain, models $M_5$ to $M_8$ are from the NLP domain. 
While these models are not publicly available, the logic of the input pipelines uses standard preprocessing operations for vision and NLP domains~\cite{shorten2019survey,shorten2021text}, similar to the open-source  input pipelines available in TF Model Garden~\cite{modelgarden}.
Models $M_1$, $M_2$ and $M_3$ are input-bound with colocated data preprocessing on our machines (and hence can benefit from horizontal scale-out), whereas the other models are not input-bound. %
We also include a canonical open-source model in our evaluation: ResNet50~\cite{resnet} trained on the ImageNet~\cite{imagenet} dataset with its default input pipeline augmented with AutoAugment~\cite{autoaugment}. 
We use the open-source implementations offered in TF Model Garden for both models and input pipelines~\cite{modelgarden}.

We use $M_1, M_2$, $M_3$, and the open-source ResNet50 model to evaluate benefits of horizontal scale-out, as these are examples of models that are input-bound with colocated data processing on our hardware and can benefit from this feature of \tfdataservice. We use $M_4$ to evaluate ephemeral data sharing, as it is a model often trained internally with the same input data pipeline for hyperparameter tuning and hence benefits from the ephemeral data sharing feature. We evaluate coordinated reads on the NLP models ($M_5$ to $M_8$) as they are not input-bound, but suffer from stalls in distributed training due to variations in input data sizes that lead to stragglers, which is what the coordinated reads feature is designed to alleviate. While we demonstrate the benefits of \tfdataservice features on particular categories of models, benefits can extend beyond the types of models we sample in this study.

\paragraph{Hardware} %
For experiments with production models, each client machine is equipped with TPU v4 accelerators~\cite{jouppi2023tpu}, two 240 core AMD EPYC 7B12 processors, 32GB of accelerator memory, 400 GB of DRAM memory and 2TB of SSD storage~\cite{mlperfconfig}, same as for Google's MLPerf v2 benchmark submissions~\cite{mlperfconfig}. The hardware configuration of workers and the dispatcher can vary for each model, as they are deployed on multi-tenant machines in the fleet, however they resemble general-purpose virtual machines available in public cloud-providers~\cite{google-cloud-pricing,aws-ec2-config}. Source data is stored in and read from Colossus, Google's internal distributed file system~\cite{colossus}. 

For externally reproducible experiments with open-source workloads, we use TPU v2-8 VMs from Google Cloud, equipped with 96 vCPUs and 335 GB or RAM~\cite{google-cloud-tpu-config}. Source data is stored in and read from GCS~\cite{google-cloud-storage}. The dispatcher and workers run on \texttt{n2-standard-8} VMs. Unless specified otherwise, storage, preprocessing and ML compute are all colocated in the same geographical region. Additionally, unless specified otherwise, neither the storage layer nor the network bandwidth are a bottleneck. 

\paragraph{Software Frameworks} We use Tensorflow~\cite{tensorflow} to define and execute ML training computations, though \tfdataservice can also be used with other frameworks like JAX~\cite{jax} and PyTorch~\cite{pytorch}. We deploy experiments with production workloads using Borg~\cite{verma2015borg,tirmazi2020borg}, Google's internal cluster management system, which autoscales resources with Autopilot~\cite{rzadca2020autopilot}. For open-source experiments, we orchestrate the \tfdataservice deployment with Kubernetes~\cite{kubernetes} instead of Borg and manually tune the number of workers. We disable compression between workers and clients as it would require unnecessary CPU cycles and bandwidth is not a bottleneck.

\paragraph{Baselines} For the colocated baseline, we measure training throughput (in batches per second) using \tfdata (without service) for input data preprocessing. \tfdata provides state-of-the-art colocated data preprocessing performance~\cite{tfdata}. 

We also show an \textit{ideal} training throughput baseline for input-bound jobs. We obtain the ideal throughput by training the model with a \inline{take(1).cache().repeat()} operation at the end of the input pipeline. This command retrieves the first element produced by the input pipeline, caches it, and returns it without any recomputation whenever the model requires a new batch. Hence, this baseline represents training time with an infinitely fast input pipeline. To avoid any potential network overheads, we measure ideal training time using \tfdata without service. %

We cannot compare to Meta's disaggregated data preprocessing service, DPP~\cite{dpp}, as it is a closed-source proprietary system. To the best of our knowledge, while DPP does disaggregate and scale out data preprocessing, it does not offer features such as ephemeral data sharing or coordinated reads. %

\paragraph{Metrics} We measure training throughput in batches per second as a sum across all accelerators and report speedups for each workload with versus without \tfdataservice. We quantify \tfdataservice's impact on cost using the cost model described below. To ensure disaggregation does not affect model quality, despite relaxed visitation guarantees, we compared the loss and accuracy of the trained models. We verified that the models achieved approximately the same loss and accuracy with and without \tfdataservice. %

\paragraph{Cost Model} We compute the cost $\mathcal{C}$ of an ML job using the following formula:

\begin{equation}
\begin{split}
    \mathcal{C} =&~t (\mathcal{C}_{CPU}(n_W \cdot \overline{CPU}^W_{u} + n_T \cdot CPU^T_a) \\
    & + \mathcal{C}_{MEM}(n_W \cdot \overline{MEM}^W_u + n_T \cdot MEM_a^T) \\
    & + \mathcal{C}_{ACC} \cdot n_T \cdot n_{ACC/T})
\end{split}
\label{eq:eval:cost}
\end{equation}

\noindent where $t$ represents the job execution time, $n_W$, $n_T$ and $n_{ACC/T}$ represent the number of \tfdataservice workers, clients and accelerators per client respectively. $\mathcal{C}_{CPU}$, $\mathcal{C}_{MEM}$ and $\mathcal{C}_{ACC}$ represent the normalized cost of one ``unit'' of CPU, memory and accelerator per unit of time respectively. $\overline{CPU}^W_{u}$ and $\overline{MEM}^W_u$ represent the average CPU and memory utilization across all \tfdataservice workers per unit of time. Finally, $CPU^T_a$ and $MEM_a^T$ represent the available CPU and memory for a client. We use CPU and memory utilization (as opposed to reservation) for \tfdataservice workers, as reserved but unused CPU and memory resources are released back to the global resource pool. In contrast, the CPU and memory of ML hosts are non-fungible, regardless of their true utilization, and are thus charged altogether. For experiments in the production cluster, we do not disclose the absolute or relative values of $\mathcal{C}_{CPU}$, $\mathcal{C}_{MEM}$, and $\mathcal{C}_{ACC}$ but they are consistent with the pricing of cloud VMs~\cite{aws-pricing,google-cloud-pricing}. For open-source experiments, we use the prices of Google Cloud VMs in the month of June 2023 for the region \texttt{us-central1}: 4.5\$/h for a TPU v2-8 VM, and 0.08\$/h for a \texttt{n2-standard-8} VM. The formula allows us compare the cost of different data preprocessing setups in our experiments.

We do not explicitly account for network bandwidth usage in the cost model. \tfdataservice is strongly recommended to run in the same datacenter as the ML accelerator clients, since intra-datacenter network bandwidth is cheap and abundant. For example, on Google Cloud, intra-zone network bandwidth is free, meaning it is already included in VM pricing~\cite{gcp-network-pricing}. Other cloud providers have similar policies.

\begin{figure*}[t]
  \centering
  \begin{subfigure}[]{0.45\textwidth}
    \includegraphics[trim = 0.4cm 0.4cm 0.3cm 0.3cm,clip=true, width=\linewidth]{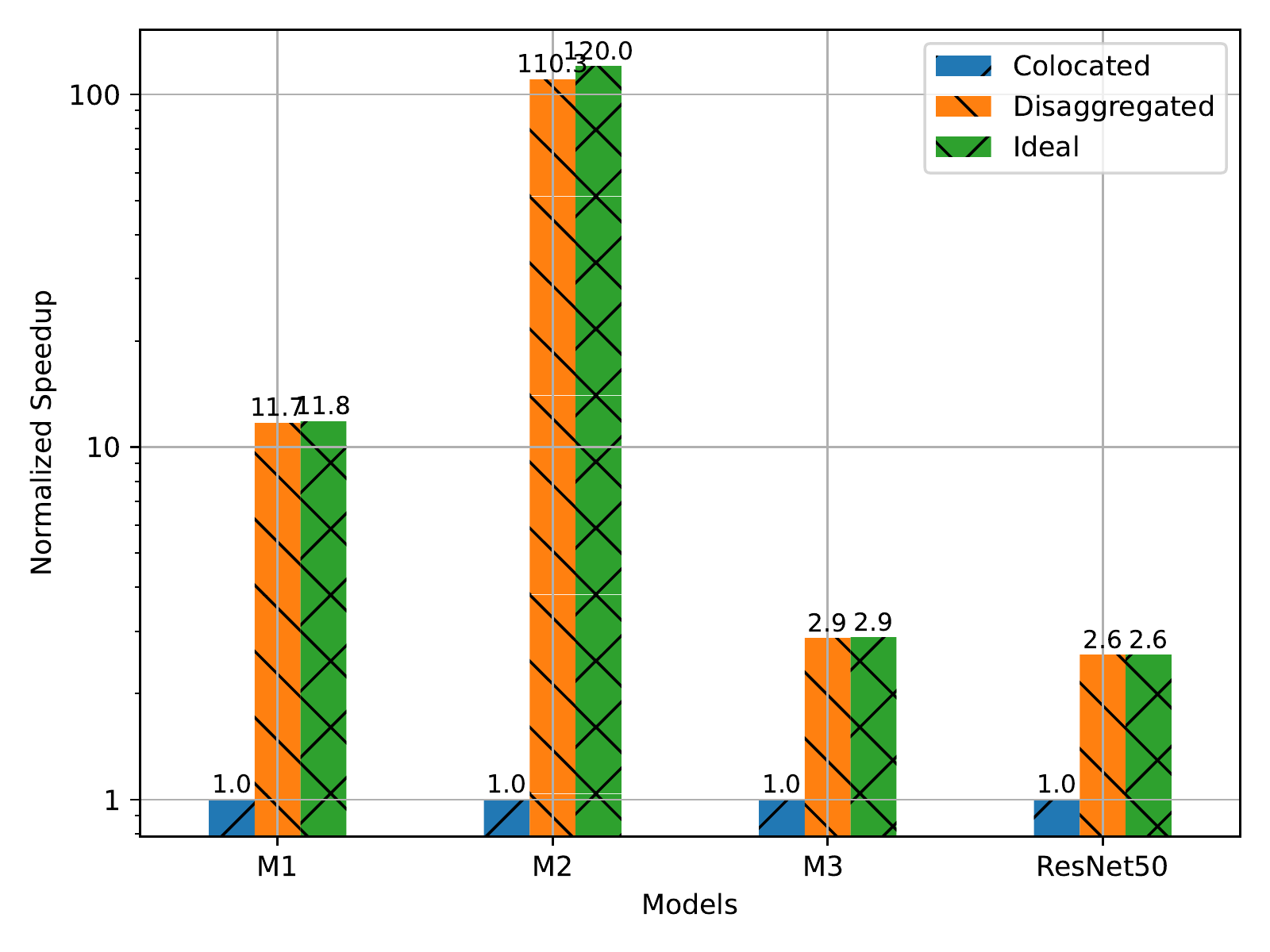}
    \caption{Job speedups}
    \label{fig:eval:distributed-preprocessing:time-savings}
  \end{subfigure}
  \hfill
  \begin{subfigure}[]{0.45\textwidth}
    \includegraphics[trim = 0.4cm 0.4cm 0.3cm 0.3cm,clip=true, width=\linewidth]{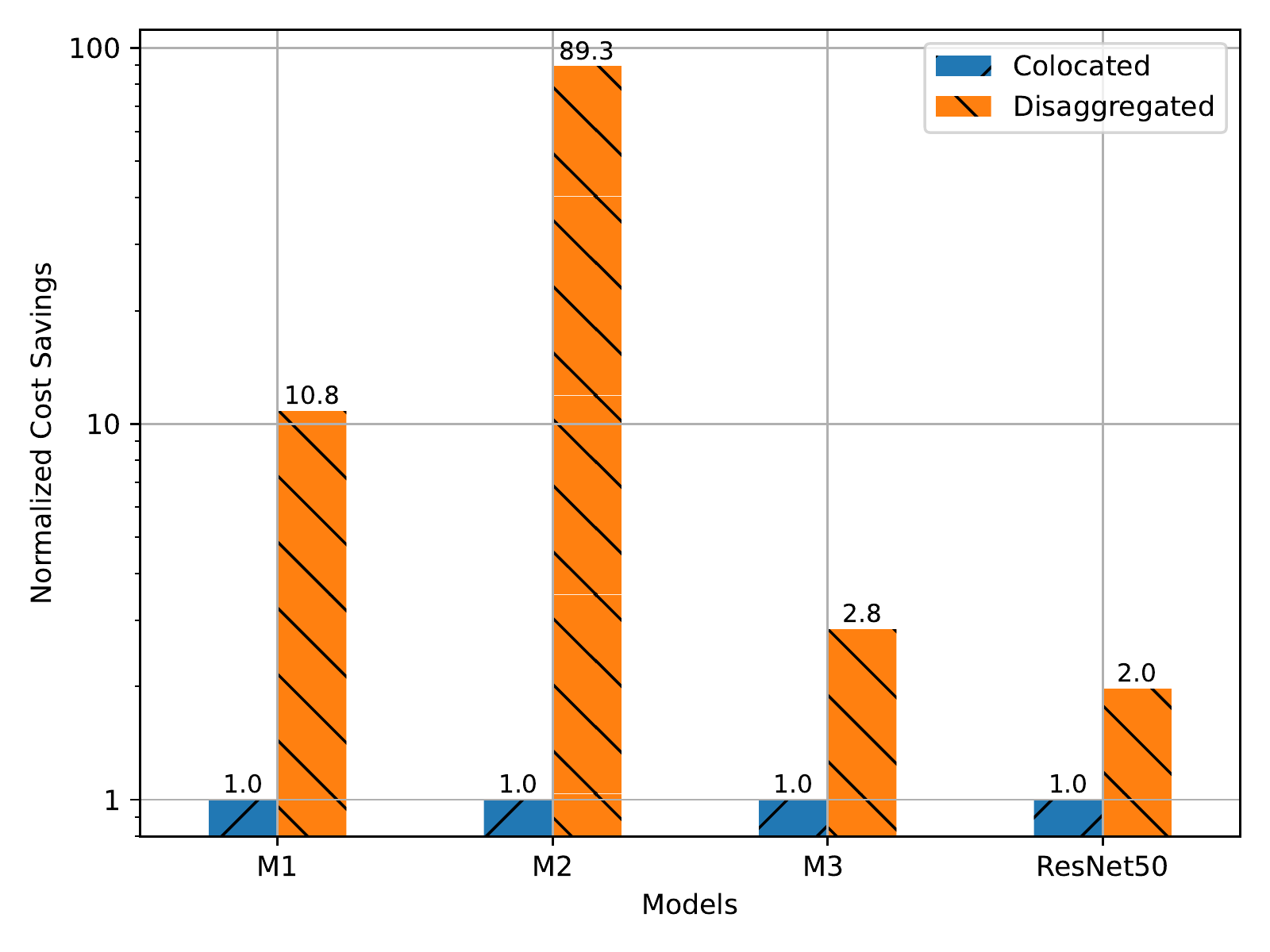}
    \caption{Job cost savings}
    \label{fig:eval:distributed-preprocessing:cost-savings}
  \end{subfigure}
  \vspace{-5pt}
  \caption{End-to-end job time speedups and cost reductions with disaggregated \tfdataservice.}
  \label{fig:eval:distributed-preprocessing}
\end{figure*}

\subsection{Horizontal Scale-Out}
\label{sec:eval:distributed-preprocessing}

To demonstrate the benefits of \tfdataservice for input-bound ML jobs, we compare training throughput and cost for models $M_1$, $M_2$, $M_3$, and ResNet50 with and without \tfdataservice. We use 32, 8, 16, and 1 accelerators to train the models, respectively. When trained without \tfdataservice, models $M_1$ and $M_2$ and ResNet50 fully utilize locally available CPU resources for data ingestion and processing, while $M_3$ makes partial use of the locally available CPU resources. We present this mix of models to demonstrate that \tfdataservice can address both hardware and software input bottlenecks.

\paragraph{Speedup}
Figure~\ref{fig:eval:distributed-preprocessing:time-savings} shows training throughput speedup with horizontal scale-out enabled by disaggregation in \tfdataservice. We observe speedups of 11.7$\times$ (from 0.55 batches/s to 6.47 batches/s) for $M_1$, 110.3$\times$ (from 4.7 batches/s to 518.4 batches/s) for $M_2$, 2.9$\times$ (from 22.2 batches/s to 63.8 batches/s) for $M_3$, and 2.57$\times$ (from 1.75 batches/s to 4.5 batches/s) for ResNet50. The average speedup across these jobs is \textbf{31.7$\times$}. The service was scaled to 442, 421, 128, and 16 workers for the $M_1$, $M_2$, $M_3$, and ResNet50 experiments, respectively. For models $M_1$  $M_3$, and ResNet50, \tfdataservice was able to reach the ideal speedup, while for $M_2$ it fell 8\% short. Upon closer inspection into model $M_2$'s disaggregated performance, we found that the model receives data from \tfdataservice without delay upon request, suggesting that the penalty in performance stems from hardware limitations on the side of the trainer nodes. The remarkably high ingestion rate requirements of this model (up to 580 batches per second) in concert with the need to deserialize and copy all batches received from \tfdataservice introduce resource contention on the trainer nodes, causing the model to slow down. This contrasts with the ideal speedup where no deserialization and additional data copies are required.  %

\begin{figure*}[t]
  \centering
  \begin{subfigure}[]{0.47\textwidth}
    \includegraphics[width=\linewidth]{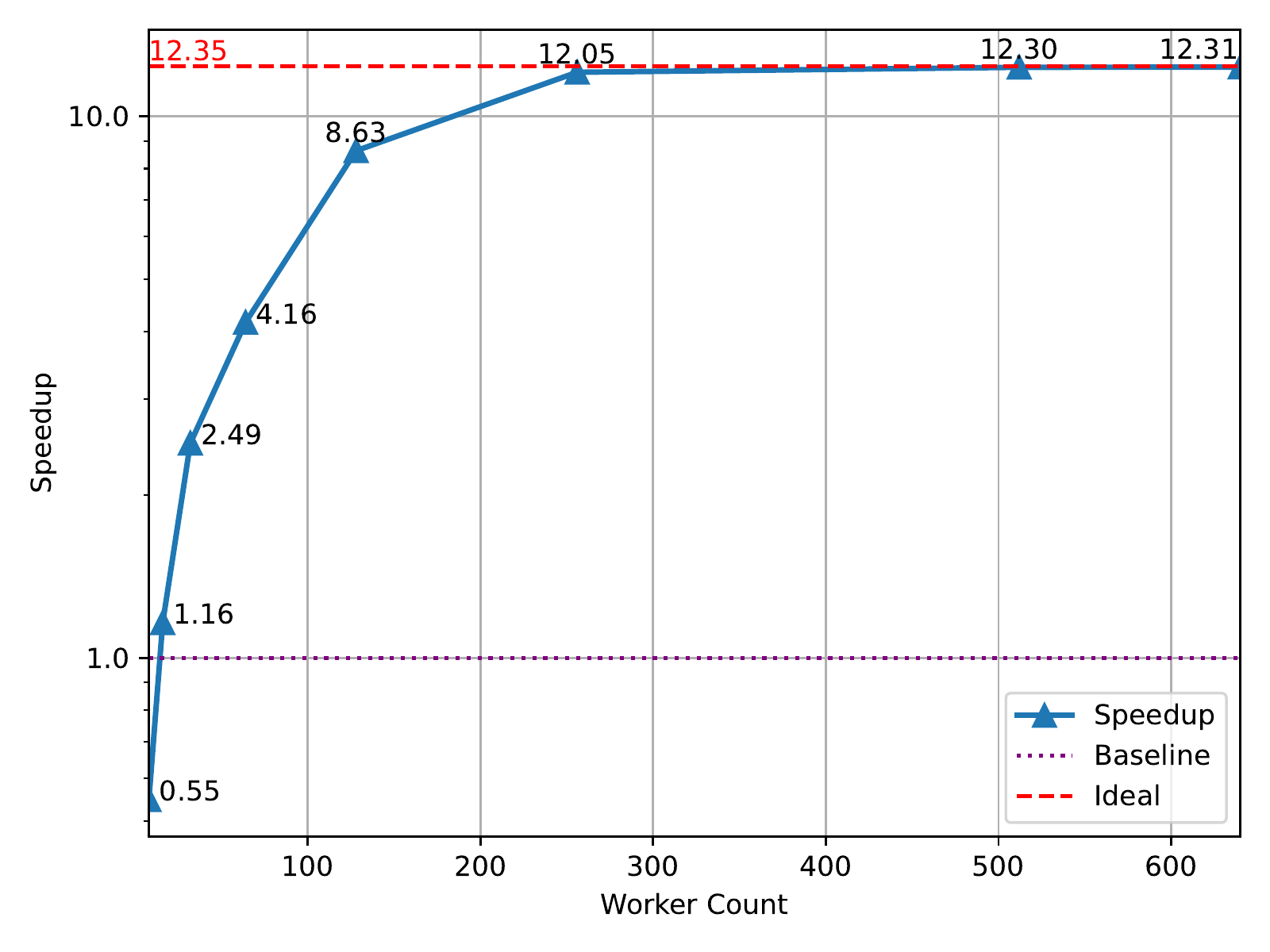}
    \caption{Job time speedup}
    \label{fig:eval:disaggregation-impact:speedup}
  \end{subfigure}
  \hfill
  \begin{subfigure}[]{0.47\textwidth}
    \includegraphics[width=\linewidth]{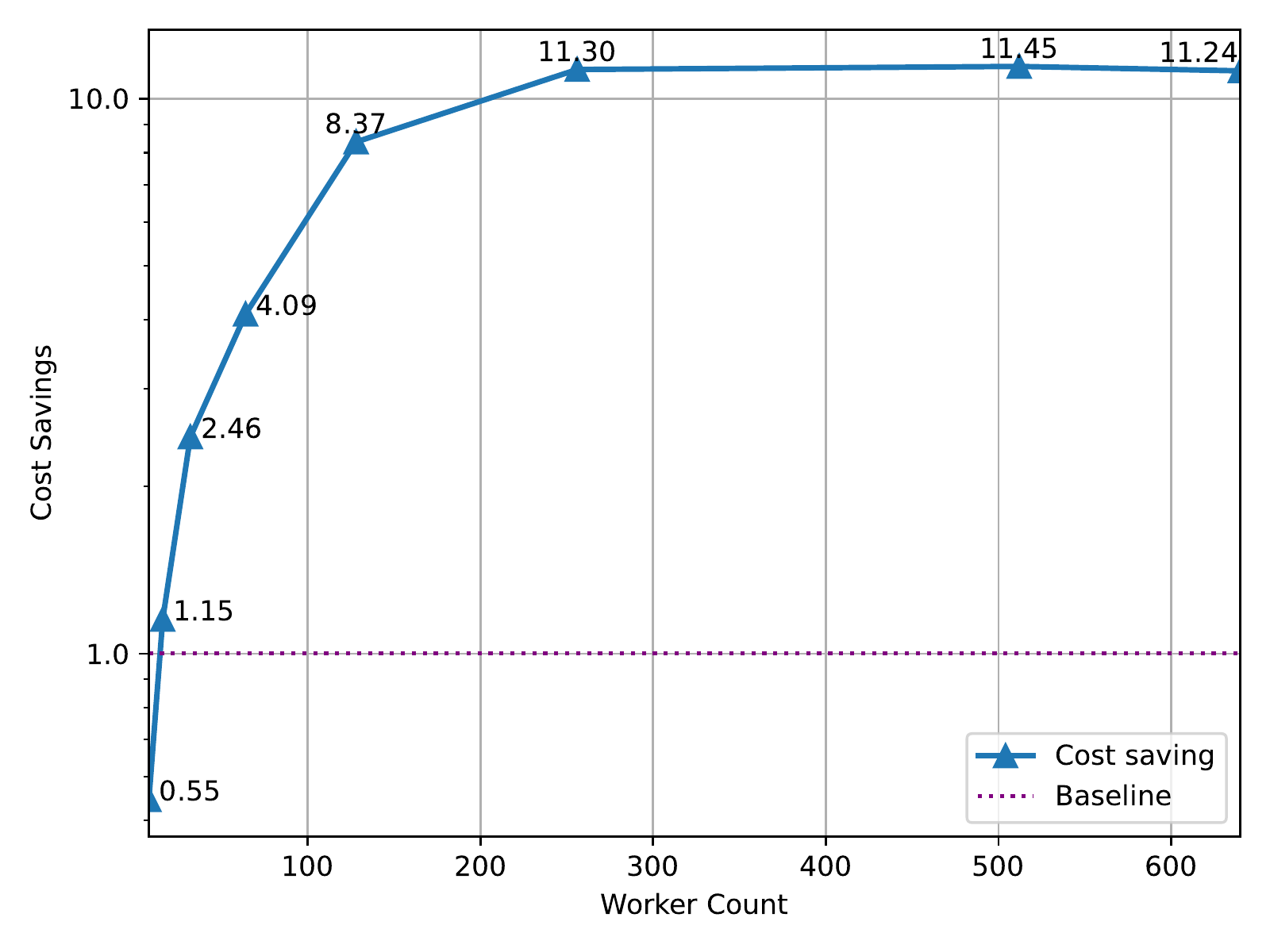}
    \caption{Job cost savings}
    \label{fig:eval:disaggregation-impact:cost}
  \end{subfigure}
  \vspace{-5pt}
  \caption{Job time speedup and cost savings for model $M_1$ across several \tfdataservice worker counts.}
  \label{fig:eval:disaggregation-impact}
\end{figure*}

\paragraph{Cost Savings} \tfdataservice achieved the aforementioned throughput benefits by using additional resources for data preprocessing. To better understand the end-to-end cost effectiveness of horizontal-scale out, we compare the cost of model training with and without \tfdataservice in Figure~\ref{fig:eval:distributed-preprocessing:cost-savings}. Notably, the cost savings are not much less than the speedups: 10.8$\times$, 89.3$\times$, 2.8$\times$, and 1.97$\times$ for models $M_1$, $M_2$, $M_3$, and ResNet50, respectively. The average cost saving across these jobs is \textbf{26.2$\times$}. The cost savings are a result of jobs completing faster and hence using expensive ML accelerator resources for less time. In the case of ResNet50, the total training cost for colocated training (i.e. only TPU VM) over 112320 steps at a batch size of 1024 is 80.2\$, while for disaggregated training with 16 workers and 1 dispatcher, the cost drops to 40.6\$ (9.4\$ for the service and 31.2\$ for the TPU VM). %

Note that our cost equation does not capture possibly the most important cost -- the opportunity cost. By allowing ML workloads to finish faster, the valuable and scarce ML hardware is released back into the general pool sooner and can be used by other ML computations. In other words, the cost savings estimated using Equation~\ref{eq:eval:cost} are likely an underestimate of the actual economical impact of using disaggregated \tfdataservice to remove input bottlenecks of ML workloads.

\paragraph{Cross-region Scenario} To further illustrate the benefits of disaggregated \tfdataservice, we evaluate a scenario where the source dataset is stored in a different georgraphical region than data preprocessing and training. This situation occurs in practice, for example due to per-region resource quota availability or data regulations. Such situations can incur large performance penalties on the job time due to increased latency caused by extra network hops between storage and compute. For this experiment, we choose $M_3$ and store its data in a different continent than the region where data preprocessing and training take place. We measure that with ``out-of-region'' data, training with local \tfdata preprocessing is 13.3$\times$ slower than ideal (compared to only 2.9$\times$ slower than ideal with the ``in-region'' scenario in Figure~\ref{fig:eval:distributed-preprocessing:time-savings}). This is due input data fetching bottlenecks. We verify that with horizontal scaling, \tfdataservice is able to reach the ideal speedup even in the ``out-of-region'' case (not plotted in the figure), by using extra workers to hide data fetching latency. %

\paragraph{Sweeping Worker Count} We train $M_1$ using a varying data preprocessing worker pool size of 8, 16, 32, 64, 128, 256, 512, and 640 workers. %
We choose the smallest worker pool size (8 workers) such that it has a comparable amount of CPU/RAM resources to the CPU/RAM resources available on training client nodes for colocated data preprocessing (remote workers have less resources than training clients hosts in our setup). We choose the largest worker pool size (640 workers) such that it easily eliminates the input bottleneck.  

Figures~\ref{fig:eval:disaggregation-impact:speedup} and~\ref{fig:eval:disaggregation-impact:cost}  plot the training time speedup and job cost savings for each configuration respectively. We normalize to the training time and cost of the colocated case. The dotted horizontal line indicates the ideal job time speedup. %

With 8 workers, training time and cost is 83\% higher than with colocated processing time (0.3 batches/s versus 0.55 batches/s). This is because CPU/RAM resources are not only used for data preprocessing computations, but also for RPC processing to send and receive data over the network and auxiliary tasks such as serialization and deserialization.
When we add 8 more workers, bringing the total to 16 workers, training throughput with \tfdataservice increases to 0.64 batches/s, which is 1.14$\times$ higher than the colocated  baseline. At 64 workers, we reach a speedup of 4.1$\times$ (2.3 batches/s) and at 128 workers, we see a 8.6$\times$ speedup (4.77 batches/s) compared to the colocated case. %
With 512 workers, \tfdataservice reaches the ideal training time, achieving a training time speedup of 12.3$\times$ and corresponding cost savings of 11.4$\times$ compared to the colocated baseline. The job cost increases marginally with 640 workers, due to the superfluous 128 \tfdataservice workers, however the job time remains unaffected. Over-provisioning data preprocessing resources increases cost but has minimal impact on job time. Our observations are consistent for the other models (not plotted). %

\paragraph{Network Impact} \review{1}{In our experiments, we deploy workers in the same region as client nodes. In such deployments, the available bandwidth is never below the client ingestion rates, hence we can achieve model-bound training time. In situations where the available network bandwidth bewteen workers and client is below the clients' ingestion rate requirements, disaggregating data preprocessing increases training time. In extreme scenarios, where the available bandwidth is below the colocated preprocessing throughput, disaggregation is strongly discouraged, as the colocated deployment would achieve better performance at lower costs.}

\review{1}{We find that network latency is less of a concern for disaggregated data preprocessing, as we can hide high network latency by using more workers. For example, we observe in our "out-of-region" experiment that the higher the network latency between workers and clients, the higher the number of workers required to hide it. This means that higher network latency results in higher overall cost, as each remote worker contributes to cost. However, the dominant cost component in most deployments is still the TPU hardware. In extreme cases, where latency is exceptionally high, remote worker costs can outweigh the benefits of removing the input bottleneck. In these situations, the practitioner will need to decide if they care more about optimizing for training time or total cost.}

\textbf{\textit{Takeaway.}}
\tfdataservice enables horizontally scaling out data processing to remote CPU servers, which effectively eliminates input data stalls for jobs that are input-bound with colocated data preprocessing. Even when accounting for the cost of the extra remote CPU servers in the service deployment, we observe significant overall cost savings for training jobs (26.2$\times$ on average) with disaggregated data preprocessing due to substantial speedups (31.7$\times$ on average) that come from ensuring ML accelerators do not stall for data. The speedups translate into proportional cost savings as jobs consume expensive ML accelerators for less time.

\subsection{Ephemeral Data Sharing}

We evaluate the performance of ephemeral data sharing on a set of hyperparameter tuning jobs across three different scenarios: (A) all hyperparameter tuning jobs share a single \tfdataservice deployment with data sharing enabled, (B) all hyperparameter tuning jobs share a single \tfdataservice deployment with no data sharing enabled, (C) each hyperparameter tuning job uses a dedicated \tfdataservice deployment. We evaluate data sharing using a hyperparameter tuning job, as these types of workloads most frequently employ this feature.

\begin{figure}[t]
\includegraphics[trim = 0cm 0.4cm 0cm 0cm,clip=true,width=\linewidth]{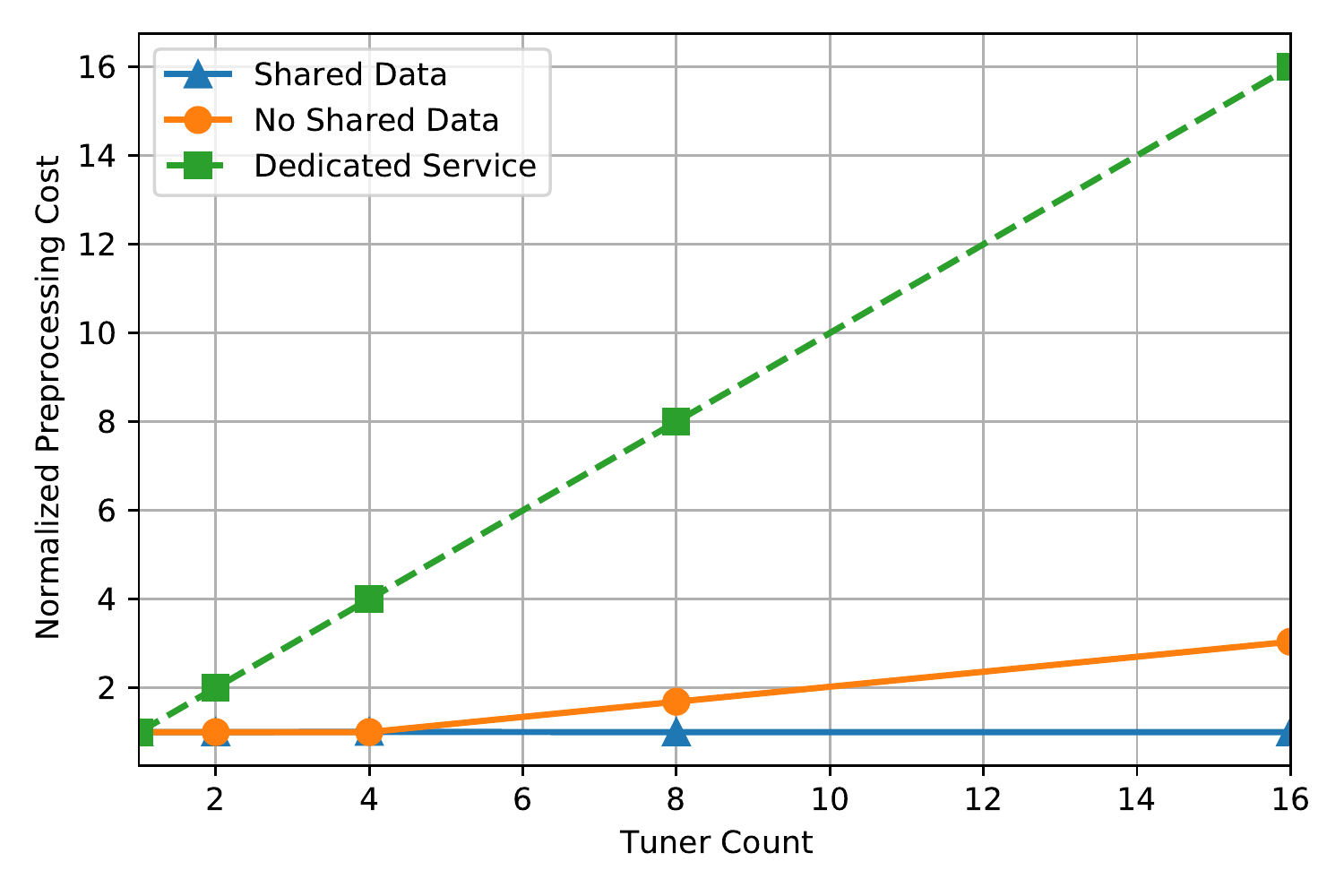}
\caption{Preprocessing costs in various deployment modes for hyperparameter tuning jobs.}
\label{fig:eval:data-sharing-lrv}
\vspace{-10pt}
\end{figure}

For modes (A) and (B), we deploy \tfdataservice with 128 workers, whereas for mode (C) each individual \tfdataservice deployment has 128 workers. We vary the number of $M_4$ model hyperparameter tuning jobs over the set $\{1, 2, 4, 8, 16\}$. A single tuning job for this model makes use of 16 accelerators. The model is not input-bound when there are at least 128 \tfdataservice workers: at this point, it reaches its ideal training throughput of 1.92 batches/s.

Our results show that mode (A) of employing a shared \tfdataservice incurs no performance overhead on the job time regardless of the number of tuning jobs. We test that this holds up to 64 tuning jobs (we did not test beyond this, since it is sufficient for our use case). With mode (B), which employs a single \tfdataservice with no data sharing, the preprocessing resources are sufficient to support up to four tuning jobs. For more concurrent jobs, the job time gradually increases by 1.75$\times$ for 8 tuning jobs (from 1.92 batches/s to 1.09 batches/s) and to 3$\times$ for 16 tuning jobs (0.64 batches/s). The longer job time maps directly to increased cost. For mode (C), which uses several dedicated \tfdataservice deployments, the job time is not affected, however the processing cost grows linearly with the number of tuning jobs. Figure~\ref{fig:eval:data-sharing-lrv} presents the normalized preprocessing cost of each of the presented modes. We normalize relative to a training run of the model without any hyperparameter tuning and with a dedicated \tfdataservice deployment. 

Besides the job time and cost benefits, \tfdataservice with data sharing helps reduce storage and network bandwidth usage by keeping the number of connections to and the number of bytes read from storage layer constant with respect to the number of tuning jobs. This contrasts with the other two modes, in which the number of connections and bytes read scales linearly in the number of tuning jobs. This feature also helps free up significant CPU and memory resources that can be used immediately by other jobs being scheduled. This reduces scheduling times and increases cluster efficiency.

\textbf{\textit{Takeaway.}}
Ephemeral data sharing in the context of concurrent jobs with identical input data pipelines (e.g. hyperparameter tuning or model search training workflows) can linearly save preprocessing resources (tested up to 64 $M_4$ jobs) without affecting the average end-to-end time of the jobs. This feature helps reduce network and storage bandwidth usage by avoiding redundant data transfers, and saves CPU/MEM resources, improving cluster efficiency. %

\subsection{Coordinated Reads}\label{eval:coordinated-reads}

To quantify the benefits of the coordinated reads feature, we evaluate the performance of models $M_5$, $M_6$, $M_7$, and $M_8$. Since none of these models are input-bound with colocated data preprocessing, all observed performance benefits stem from the ability of the coordinated reads feature to reduce imbalance of input data batch sizes fed to clients each training step. We deploy the models on 64, 8, 64 and 4 accelerators, respectively. The maximum sequence length for each of the models is 512. We use \tfdata with no service for the baseline training time. We deploy \tfdataservice with a similar amount of preprocessing resources as colocated data preprocessing to measure the coordinated reads training time. The service deployment has 4, 1, 4, and 1 worker for models $M_5$ through $M_8$, respectively. The bucket boundaries are defined at every multiple of 64 (i.e. $(0, 64]$, $(64, 128]$, etc.) for $M_5$ and $M_7$ and 128 for $M_6$ and $M_8$, up to the maximal sequence length. 

\begin{figure}[t]
\includegraphics[trim = 0cm 0.4cm 0cm 0cm,clip=true,width=0.9\linewidth]{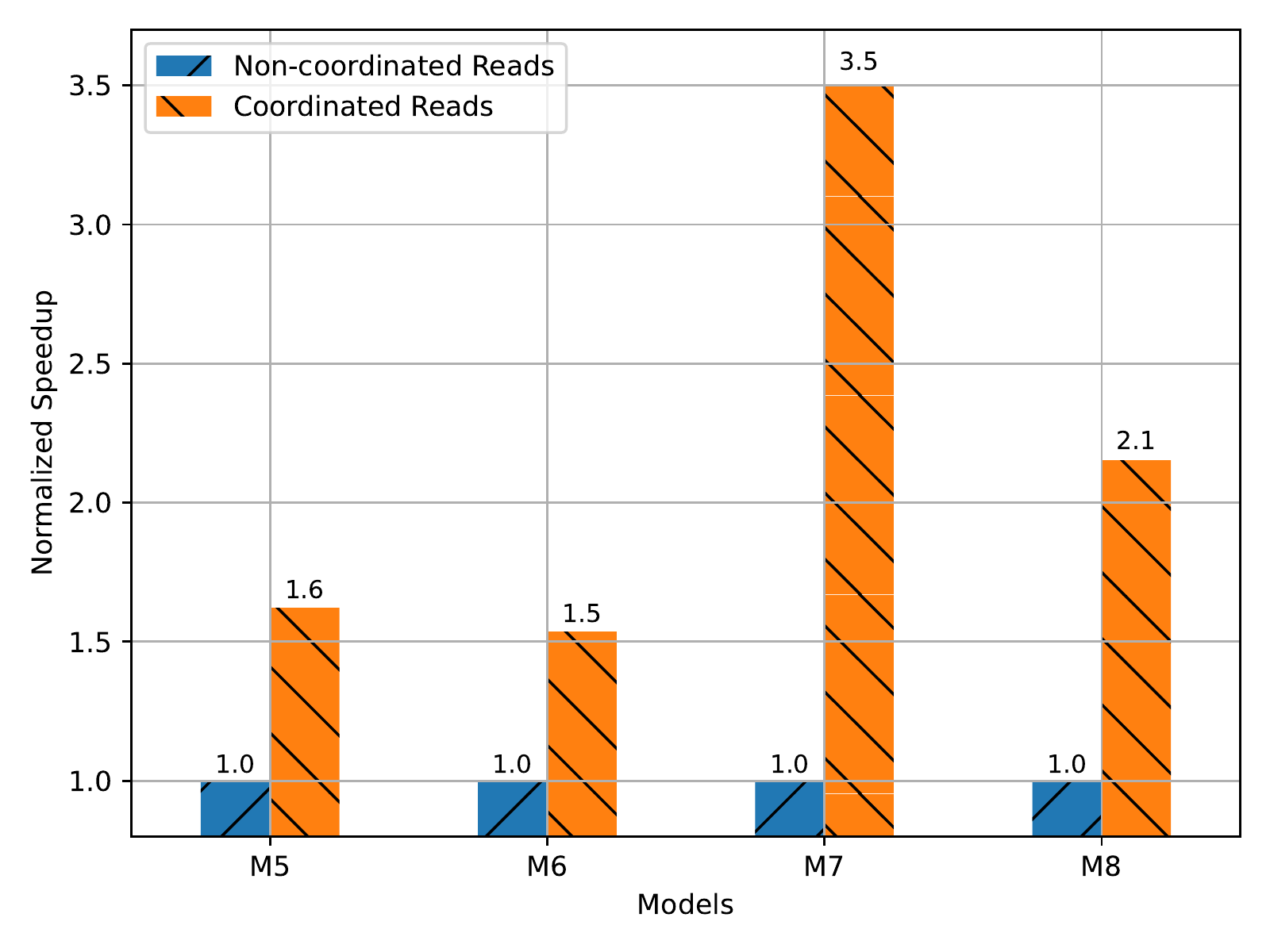}
\caption{Job time speedup in NLP production models with dynamic sequence length support.}
\label{fig:eval:coordinated-reads}
\vspace{-10pt}
\end{figure}

Figure~\ref{fig:eval:coordinated-reads} shows the training time speedups. With \tfdataservice coordinated reads, we observe speedups ranging from 1.5$\times$ to 3.5$\times$, with an average of 2.2$\times$ across the four jobs. More specifically the jobs performance improves by: ($M_5$) 1.62$\times$ (3.18 batches/s to 5.15 batches/s) ($M_6$) 1.53$\times$ (11.9 batches/s to 18.3 batches/s) ($M_7$) 3.5$\times$ (2 batches/s to 7 batches/s) ($M_8$) 2.15$\times$ (5.9 batches/s to 12.7 batches/s). These speedups correspond to equivalent job cost savings. 

\textbf{\textit{Takeaway.}}
Coordinated reads helps ensure that all batches, at every training step, have similar size and employ minimal padding, leading to uniform processing time across the training clients at every step. Coordinated reads help synchronization overheads due to stragglers, realizing speedups of 2.2$\times$ on average for the four NLP jobs in our evaluation and equivalent cost savings.

\subsection{Fleetwide Usage}

We conclude our evaluation with some more general insights of  \tfdataservice usage across the Google fleet. We launched \tfdataservice internally in Q2 2020 and open-sourced it in Q3 2020~\cite{tfdataservice-doc}. Its adoption has since been growing steadily both internally and externally (e.g. for Cloud TPU workloads).

\begin{figure*}[htb]
  \centering
  \begin{subfigure}[t]{0.46\textwidth}
    \centering
    \includegraphics[width=\textwidth]{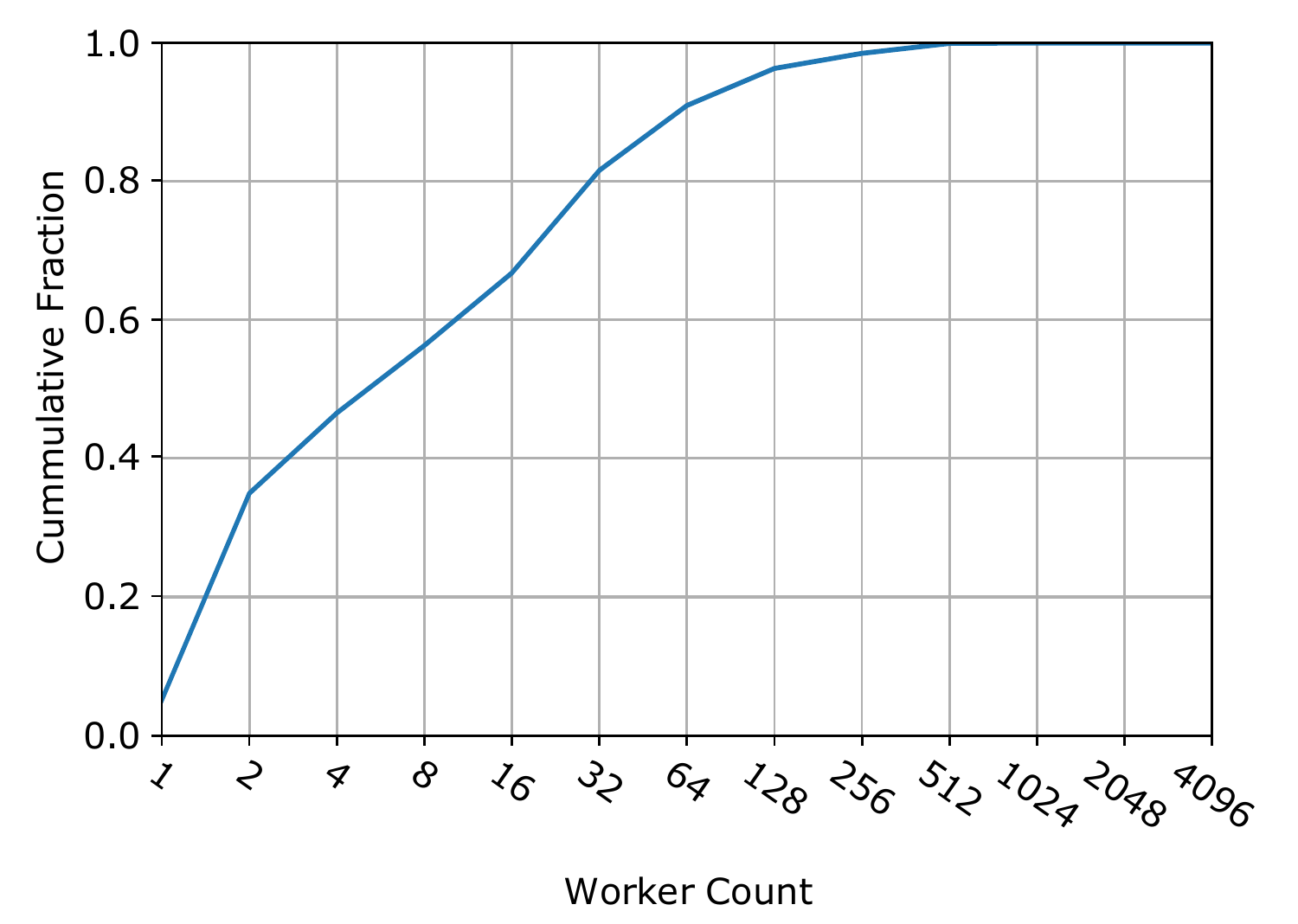}
    \caption{CDF of worker counts in production jobs.}
    \label{fig:num-jobs-by-workers}
  \end{subfigure}
  \hfill
  \begin{subfigure}[t]{0.46\textwidth}
    \centering
    \includegraphics[width=\textwidth]{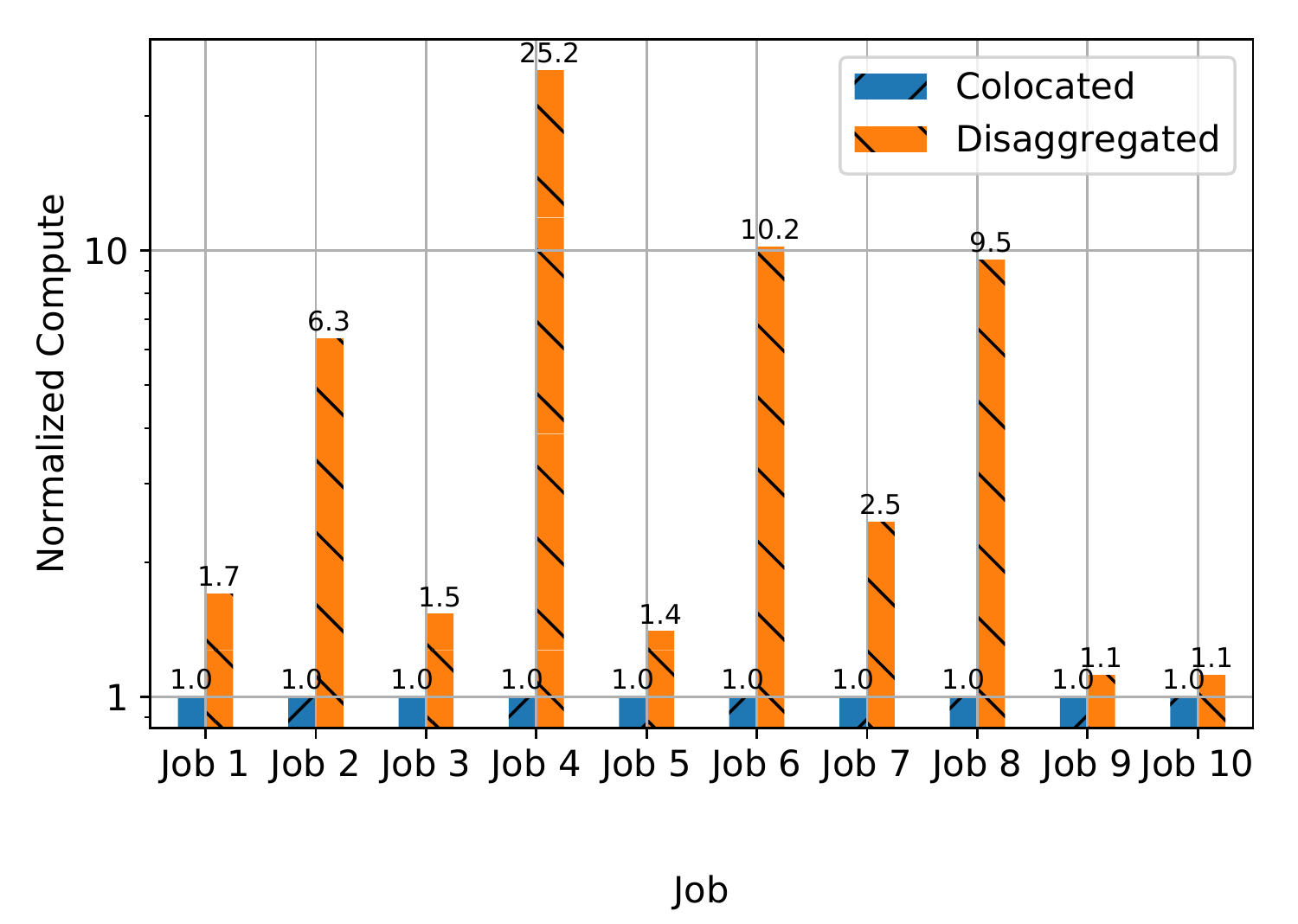}
    \caption{Client node CPU limit and tf.data worker CPU usage.}
    \label{fig:fleetwide-cpu-usage}
  \end{subfigure}
  \caption{\tfdataservice fleetwide usage.}
\end{figure*}

\paragraph{Deployment sizes} Figure~\ref{fig:num-jobs-by-workers} shows the distribution of \tfdataservice deployment sizes (i.e. the number of data processing workers) for internal ML workloads over the last year. While most training jobs deploy between 2 and 32 workers, the largest model uses more than 5K workers. %
In our fleet, \tfdataservice workers are heterogeneous (i.e., they have different amounts of CPU and RAM resources) 
as Borg schedules workers on multi-tenant machines with fungible resources.
The optimal number of data processing workers and their size depends on the model, as shown by the wide distribution in  Figure~\ref{fig:num-jobs-by-workers}.

\paragraph{Scale-out CPU usage} We  compare the total CPU usage with versus without \tfdataservice for the top 10 most CPU-intensive \tfdataservice jobs in the fleet. Figure~\ref{fig:fleetwide-cpu-usage} shows the CPU usage for jobs with disaggregated data preprocessing, normalized to their CPU usage with colocated data preprocessing.
The \tfdataservice workers use up to $25\times$ more CPU cores than available locally on ML hosts with colocated preprocessing. Without \tfdataservice, data processing would incur large input bottlenecks, slowing down the ML computation and limiting utilization of valuable ML hardware.

\section{Discussion}
\label{sec:experience}

We discuss insights from the internal adoption of \tfdataservice and the implications ML disaggregation has on future research and preprocessing dynamics. %

\paragraph{Target workloads} \review{4}{ML jobs can benefit from our work on disaggregation in three different ways: (1) eliminating input bottlenecks via horizontal-scale out (2) avoiding recomputation via ephemeral data sharing and (3) avoiding stragglers in distributed training via coordinated reads. We expect ML jobs in the image, video and audio domains to benefit the most from (1) as their input pipelines often employ complex transformations that manipulate large volumes of data, making them more likely to be input-bound. %
NLP jobs often have less compute intensive input pipelines and are thus less likely to benefit from (1). All types of ML workloads can benefit from (2) when their execution overlaps in time and they have the same input pipeline. For instance, this scenario often occurs for hyperparameter tuning jobs. Finally, NLP models frequently benefit from (3), as their input data has irregular sequence length, which can lead to straggler trainer nodes in synchronous distributed training. We are exploring how coordinated reads may also benefit other ML domains with variable input sizes (e.g. different image or audio resolutions). We note that jobs can benefit from a combination of the features brought forward by our work, and are not bound to one. For instance, a set of hyperparameter tuning NLP jobs could benefit from tf.data service to reduce redundant computation via ephemeral data sharing and eliminate straggler clients via coordinated reads.} 

\paragraph{Implications for ML accelerator host design} %
The current generation of Cloud TPU hosts provide 96 CPU cores and 335 GB of RAM per TPU accelerator, which provides significant local resources for data processing. In production, we find that approximately 20\% of accelerator usage benefits from distributed data processing since it needs even more CPU and/or memory resources than what is available on TPU hosts to avoid input data processing bottlenecks. Why not build even beefier accelerator hosts to satisfy the needs of the most data processing intensive jobs and colocate other workloads on any spare CPU/memory of hosts with less data processing intensive ML jobs? As shown in Figure~\ref{fig:intro:bursty}, sharing the host resources of ML accelerator nodes is challenging because CPU usage tends to be extremely bursty. CPU utilization oscillates from high when preprocessing a batch of data to low while the accelerator computes the model training step on this batch, making ML data processing a challenging workload to colocate with other jobs. Thus, to avoid performance interference, TPU hosts are dedicated machines -- they are not shared between users like other machines in Google's cluster~\cite{verma2015borg}.

With the support for disaggregated data processing provided by tf.data service, it is worth exploring the design of ML accelerator nodes with wimpier resources to reduce costs. For example, the job in Figure~\ref{fig:intro:bursty} under-utilizes host memory. However, other jobs that will use the same server configuration may need more CPU and/or memory. 
Disaggregation has the potential to ensure all ML jobs are provisioned with the right CPU/MEM to accelerator ratio. Just as disaggregating data storage from computation in cloud applications has led to new hardware trends such as SmartNIC-attached storage device arrays~\cite{stingray, smartNICjbofs}, it is interesting to explore hardware deployments that completely separate ML accelerators from the CPU and RAM resources used for traditional data processing. How to design SmartNIC-attached ML accelerator arrays and to what extent can such hardware improve the performance and cost of ML jobs remain open questions. 

\paragraph{ML data processing as a multi-tenant service} 
While tf.data service is currently deployed per job or set of jobs submitted by a user, we are exploring how to run tf.data service as a long-lived, multi-tenant service that is shared across an organization. Such a deployment would provide a global view of all jobs executed within an organization, which can help optimize job performance and cluster resources efficiency. This setup goes hand in hand with novel policy development which exploits global information to remove cumbersome decisions from practitioners, helping both preprocessing performance and model quality. The advantage of features that exploit the organization's view is that they can help make globally optimal decisions, in contrast to practitioners which have a limited view of only their jobs. For example, Cachew~\cite{graur2022cachew} is a system built on top of \tfdataservice that caches common input data pipelines across jobs from different tenants to optimize training throughput. Cachew also autoscales the number of remote workers for each job to maximize throughput-per-dollar. We see further opportunities to leverage global knowledge across jobs: for dataset discovery~\cite{neural-server-scalable, neural-data-server} and automatic input pipeline augmentations~\cite{autoaugment,randaugment}. Sharing preprocessing at a finer granularity within the input pipeline (similar to shared query execution~\cite{giannikis2014shared, harizopoulos2005qpipe, marroquin2018pay}) and combining this with semantic transformations of the input pipeline (similar to semantic query optimization~\cite{logic_based_approach_semantic_query_opt,  hirzel2014catalog, semantic_query_opt}) are other promising directions. Disaggregation enables a rich design space for resource and data management policies, to optimize preprocessing throughput and hardware utilization in concert with training dynamics.

\section{Conclusion}
\label{sec:conclusion}
 
We presented \tfdataservice and showed that  disaggregating ML input data preprocessing from ML training allows us to eliminate input data stalls with horizontal scale-out, achieving up to 31.7$\times$ speedups and 26.2$\times$ cost savings, on average. We also showed that disaggregation is useful beyond horizontal scaling, as it enables other features in \tfdataservice, such as ephemeral data sharing, which improves resource efficiency by avoiding redundant computations in concurrent jobs with no performance loss, and coordinated reads, which helps eliminate straggler clients in synchronous distributed training (2.2$\times$ average speedup for NLP jobs). %
Our work motivates future research in ML hardware and data processing service design to further improve cluster efficiency and model quality. %

\begin{acks}
This work represents a combined effort of a large group of people. First, we would like to acknowledge Battulga Bayarsaikhan, Emanuel Taropa, Rohan Anil, and Ryan Doherty for their work on DataGuzzler~\cite{data-guzzler}, a precursor to tf.data service. Next, we would like to thank everyone who contributed to the development of tf.data service, most notably Derek Murray and Rohan Jain, and everyone who provided feedback to help us improve tf.data service, including Aleksandr Zaks, Aniruddh Nath, Deping Xie, Hao Wu, Huan Gui, Josh Cai, Lin Yuan, Mahesh Sathiamoorthy, Ruoxin Sang, Yingpei Wang, and Zhaoqi Leng. Last but not least, we would like to thank Abhishek Gupta, Carrie Grimes Bostock, and Steven Hand for reviewing the paper and helping us improve it.
\end{acks}

\bibliographystyle{ACM-Reference-Format}
\bibliography{main.bib}

%%% -*-BibTeX-*-
%%% Do NOT edit. File created by BibTeX with style
%%% ACM-Reference-Format-Journals [18-Jan-2012].

\begin{thebibliography}{77}

%%% ====================================================================
%%% NOTE TO THE USER: you can override these defaults by providing
%%% customized versions of any of these macros before the \bibliography
%%% command.  Each of them MUST provide its own final punctuation,
%%% except for \shownote{}, \showDOI{}, and \showURL{}.  The latter two
%%% do not use final punctuation, in order to avoid confusing it with
%%% the Web address.
%%%
%%% To suppress output of a particular field, define its macro to expand
%%% to an empty string, or better, \unskip, like this:
%%%
%%% \newcommand{\showDOI}[1]{\unskip}   % LaTeX syntax
%%%
%%% \def \showDOI #1{\unskip}           % plain TeX syntax
%%%
%%% ====================================================================

\ifx \showCODEN    \undefined \def \showCODEN     #1{\unskip}     \fi
\ifx \showDOI      \undefined \def \showDOI       #1{#1}\fi
\ifx \showISBNx    \undefined \def \showISBNx     #1{\unskip}     \fi
\ifx \showISBNxiii \undefined \def \showISBNxiii  #1{\unskip}     \fi
\ifx \showISSN     \undefined \def \showISSN      #1{\unskip}     \fi
\ifx \showLCCN     \undefined \def \showLCCN      #1{\unskip}     \fi
\ifx \shownote     \undefined \def \shownote      #1{#1}          \fi
\ifx \showarticletitle \undefined \def \showarticletitle #1{#1}   \fi
\ifx \showURL      \undefined \def \showURL       {\relax}        \fi
% The following commands are used for tagged output and should be
% invisible to TeX
\providecommand\bibfield[2]{#2}
\providecommand\bibinfo[2]{#2}
\providecommand\natexlab[1]{#1}
\providecommand\showeprint[2][]{arXiv:#2}

\bibitem[bea(2022)]%
        {beam}
 \bibinfo{year}{2022}\natexlab{}.
\newblock \bibinfo{title}{{Apache Beam: An advanced unified programming model}}.
\newblock \bibinfo{howpublished}{\url{https://beam.apache.org/}}.
\newblock


\bibitem[flu(2022)]%
        {flume}
 \bibinfo{year}{2022}\natexlab{}.
\newblock \bibinfo{title}{{Apache Flume}}.
\newblock \bibinfo{howpublished}{\url{https://flume.apache.org/}}.
\newblock


\bibitem[Abadi et~al\mbox{.}(2016)]%
        {tensorflow}
\bibfield{author}{\bibinfo{person}{Martin Abadi}, \bibinfo{person}{Paul Barham}, \bibinfo{person}{Jianmin Chen}, \bibinfo{person}{Zhifeng Chen}, \bibinfo{person}{Andy Davis}, \bibinfo{person}{Jeffrey Dean}, \bibinfo{person}{Matthieu Devin}, \bibinfo{person}{Sanjay Ghemawat}, \bibinfo{person}{Geoffrey Irving}, \bibinfo{person}{Michael Isard}, \bibinfo{person}{Manjunath Kudlur}, \bibinfo{person}{Josh Levenberg}, \bibinfo{person}{Rajat Monga}, \bibinfo{person}{Sherry Moore}, \bibinfo{person}{Derek~G. Murray}, \bibinfo{person}{Benoit Steiner}, \bibinfo{person}{Paul Tucker}, \bibinfo{person}{Vijay Vasudevan}, \bibinfo{person}{Pete Warden}, \bibinfo{person}{Martin Wicke}, \bibinfo{person}{Yuan Yu}, {and} \bibinfo{person}{Xiaoqiang Zheng}.} \bibinfo{year}{2016}\natexlab{}.
\newblock \showarticletitle{{TensorFlow: A system for large-scale machine learning}}. In \bibinfo{booktitle}{\emph{Proc. of OSDI}}.
\newblock
\urldef\tempurl%
\url{https://www.usenix.org/system/files/conference/osdi16/osdi16-abadi.pdf}
\showURL{%
\tempurl}


\bibitem[{Amazon}(2022a)]%
        {aws-pricing}
\bibfield{author}{\bibinfo{person}{{Amazon}}.} \bibinfo{year}{2022}\natexlab{a}.
\newblock \bibinfo{title}{{Amazon EC2 Pricing}}.
\newblock \bibinfo{howpublished}{\url{https://aws.amazon.com/ec2/pricing/}}.
\newblock


\bibitem[{Amazon}(2022b)]%
        {aws-ec2-config}
\bibfield{author}{\bibinfo{person}{{Amazon}}.} \bibinfo{year}{2022}\natexlab{b}.
\newblock \bibinfo{title}{{Amazon EC2 Pricing}}.
\newblock \bibinfo{howpublished}{\url{https://aws.amazon.com/ec2/instance-types/}}.
\newblock


\bibitem[Anil et~al\mbox{.}(2021)]%
        {data-guzzler}
\bibfield{author}{\bibinfo{person}{Rohan Anil}, \bibinfo{person}{Battulga Bayarsaikhan}, \bibinfo{person}{Ryan Doherty}, {and} \bibinfo{person}{Emanuel Taropa}.} \bibinfo{year}{2021}\natexlab{}.
\newblock \bibinfo{title}{Distributed computing pipeline processing}.
\newblock \bibinfo{howpublished}{https://patents.google.com/patent/WO2021177976A1}.
\newblock


\bibitem[Bottou(2009)]%
        {bottou-curiously-fast-convergence}
\bibfield{author}{\bibinfo{person}{Leon Bottou}.} \bibinfo{year}{2009}\natexlab{}.
\newblock \showarticletitle{{Curiously Fast Convergence of some Stochastic Gradient Descent Algorithms}}. In \bibinfo{booktitle}{\emph{Proc. of the Symposium on Learning and Data Science}}.
\newblock


\bibitem[Bottou(2010)]%
        {bottou2010large}
\bibfield{author}{\bibinfo{person}{L{\'e}on Bottou}.} \bibinfo{year}{2010}\natexlab{}.
\newblock \showarticletitle{Large-scale machine learning with stochastic gradient descent}. In \bibinfo{booktitle}{\emph{Proceedings of COMPSTAT'2010: 19th International Conference on Computational StatisticsParis France, August 22-27, 2010 Keynote, Invited and Contributed Papers}}. Springer, \bibinfo{pages}{177--186}.
\newblock


\bibitem[Bradbury et~al\mbox{.}(2018)]%
        {jax}
\bibfield{author}{\bibinfo{person}{James Bradbury}, \bibinfo{person}{Roy Frostig}, \bibinfo{person}{Peter Hawkins}, \bibinfo{person}{Matthew~James Johnson}, \bibinfo{person}{Chris Leary}, \bibinfo{person}{Dougal Maclaurin}, {and} \bibinfo{person}{Skye Wanderman-Milne}.} \bibinfo{year}{2018}\natexlab{}.
\newblock \bibinfo{booktitle}{\emph{{JAX}: composable transformations of {P}ython+{N}um{P}y programs}}.
\newblock
\urldef\tempurl%
\url{http://github.com/google/jax}
\showURL{%
\tempurl}


\bibitem[Broadcom(2019)]%
        {stingray}
\bibfield{author}{\bibinfo{person}{Broadcom}.} \bibinfo{year}{2019}\natexlab{}.
\newblock \bibinfo{booktitle}{\emph{Broadcom Stingray PS250 SmartNIC}}.
\newblock
\urldef\tempurl%
\url{https://docs.broadcom.com/doc/PS250-PB}
\showURL{%
\tempurl}


\bibitem[Cao et~al\mbox{.}(2021)]%
        {neural-server-scalable}
\bibfield{author}{\bibinfo{person}{Tianshi Cao}, \bibinfo{person}{Sasha~(Alexandre) Doubov}, \bibinfo{person}{David Acuna}, {and} \bibinfo{person}{Sanja Fidler}.} \bibinfo{year}{2021}\natexlab{}.
\newblock \showarticletitle{Scalable Neural Data Server: A Data Recommender for Transfer Learning}. In \bibinfo{booktitle}{\emph{Advances in Neural Information Processing Systems}}, \bibfield{editor}{\bibinfo{person}{M.~Ranzato}, \bibinfo{person}{A.~Beygelzimer}, \bibinfo{person}{Y.~Dauphin}, \bibinfo{person}{P.S. Liang}, {and} \bibinfo{person}{J.~Wortman Vaughan}} (Eds.).
\newblock


\bibitem[Chakravarthy et~al\mbox{.}(1990)]%
        {logic_based_approach_semantic_query_opt}
\bibfield{author}{\bibinfo{person}{Upen~S. Chakravarthy}, \bibinfo{person}{John Grant}, {and} \bibinfo{person}{Jack Minker}.} \bibinfo{year}{1990}\natexlab{}.
\newblock \showarticletitle{Logic-Based Approach to Semantic Query Optimization}.
\newblock \bibinfo{journal}{\emph{ACM Trans. Database Syst.}} \bibinfo{volume}{15}, \bibinfo{number}{2} (\bibinfo{date}{jun} \bibinfo{year}{1990}), \bibinfo{pages}{162–207}.
\newblock
\showISSN{0362-5915}
\urldef\tempurl%
\url{https://doi.org/10.1145/78922.78924}
\showDOI{\tempurl}


\bibitem[Choi et~al\mbox{.}(2019)]%
        {data-echoing}
\bibfield{author}{\bibinfo{person}{Dami Choi}, \bibinfo{person}{Alexandre Passos}, \bibinfo{person}{Christopher~J. Shallue}, {and} \bibinfo{person}{George~E. Dahl}.} \bibinfo{year}{2019}\natexlab{}.
\newblock \bibinfo{title}{{Faster Neural Network Training with Data Echoing}}.
\newblock
\newblock
\showeprint[arxiv]{1907.05550}~[cs.LG]


\bibitem[Contributors(2022)]%
        {pytorch-dataloader}
\bibfield{author}{\bibinfo{person}{Torch Contributors}.} \bibinfo{year}{2022}\natexlab{}.
\newblock \bibinfo{title}{{PyTorch Docs: torch.utils.data}}.
\newblock \bibinfo{howpublished}{\url{https://pytorch.org/docs/stable/data.html}}.
\newblock


\bibitem[Cubuk et~al\mbox{.}(2019)]%
        {autoaugment}
\bibfield{author}{\bibinfo{person}{Ekin~D. Cubuk}, \bibinfo{person}{Barret Zoph}, \bibinfo{person}{Dandelion Man{\'{e}}}, \bibinfo{person}{Vijay Vasudevan}, {and} \bibinfo{person}{Quoc~V. Le}.} \bibinfo{year}{2019}\natexlab{}.
\newblock \showarticletitle{AutoAugment: Learning Augmentation Strategies From Data}. In \bibinfo{booktitle}{\emph{{IEEE} Conference on Computer Vision and Pattern Recognition, {CVPR}}}.
\newblock


\bibitem[Cubuk et~al\mbox{.}(2020)]%
        {randaugment}
\bibfield{author}{\bibinfo{person}{Ekin~Dogus Cubuk}, \bibinfo{person}{Barret Zoph}, \bibinfo{person}{Jon Shlens}, {and} \bibinfo{person}{Quoc Le}.} \bibinfo{year}{2020}\natexlab{}.
\newblock \showarticletitle{RandAugment: Practical Automated Data Augmentation with a Reduced Search Space}. In \bibinfo{booktitle}{\emph{Advances in Neural Information Processing Systems}}, \bibfield{editor}{\bibinfo{person}{H.~Larochelle}, \bibinfo{person}{M.~Ranzato}, \bibinfo{person}{R.~Hadsell}, \bibinfo{person}{M.~F. Balcan}, {and} \bibinfo{person}{H.~Lin}} (Eds.). \bibinfo{pages}{18613--18624}.
\newblock


\bibitem[Dageville et~al\mbox{.}(2016)]%
        {snowflake}
\bibfield{author}{\bibinfo{person}{Benoit Dageville}, \bibinfo{person}{Thierry Cruanes}, \bibinfo{person}{Marcin Zukowski}, \bibinfo{person}{Vadim Antonov}, \bibinfo{person}{Artin Avanes}, \bibinfo{person}{Jon Bock}, \bibinfo{person}{Jonathan Claybaugh}, \bibinfo{person}{Daniel Engovatov}, \bibinfo{person}{Martin Hentschel}, \bibinfo{person}{Jiansheng Huang}, \bibinfo{person}{Allison~W. Lee}, \bibinfo{person}{Ashish Motivala}, \bibinfo{person}{Abdul~Q. Munir}, \bibinfo{person}{Steven Pelley}, \bibinfo{person}{Peter Povinec}, \bibinfo{person}{Greg Rahn}, \bibinfo{person}{Spyridon Triantafyllis}, {and} \bibinfo{person}{Philipp Unterbrunner}.} \bibinfo{year}{2016}\natexlab{}.
\newblock \showarticletitle{The Snowflake Elastic Data Warehouse}. In \bibinfo{booktitle}{\emph{Proceedings of the 2016 International Conference on Management of Data}} \emph{(\bibinfo{series}{SIGMOD '16})}.
\newblock


\bibitem[Deng et~al\mbox{.}(2009)]%
        {imagenet}
\bibfield{author}{\bibinfo{person}{Jia Deng}, \bibinfo{person}{W. Dong}, \bibinfo{person}{R. Socher}, \bibinfo{person}{L.-J. Li}, \bibinfo{person}{K. Li}, {and} \bibinfo{person}{L. Fei-Fei}.} \bibinfo{year}{2009}\natexlab{}.
\newblock \showarticletitle{{ImageNet: A Large-Scale Hierarchical Image Database}}. In \bibinfo{booktitle}{\emph{Proc. of CVPR}}.
\newblock


\bibitem[Geiping et~al\mbox{.}(2023)]%
        {geiping2023data}
\bibfield{author}{\bibinfo{person}{Jonas Geiping}, \bibinfo{person}{Micah Goldblum}, \bibinfo{person}{Gowthami Somepalli}, \bibinfo{person}{Ravid Shwartz-Ziv}, \bibinfo{person}{Tom Goldstein}, {and} \bibinfo{person}{Andrew~Gordon Wilson}.} \bibinfo{year}{2023}\natexlab{}.
\newblock \bibinfo{title}{How Much Data Are Augmentations Worth? An Investigation into Scaling Laws, Invariance, and Implicit Regularization}.
\newblock
\newblock
\showeprint[arxiv]{2210.06441}~[cs.LG]


\bibitem[Giannikis et~al\mbox{.}(2014)]%
        {giannikis2014shared}
\bibfield{author}{\bibinfo{person}{Georgios Giannikis}, \bibinfo{person}{Darko Makreshanski}, \bibinfo{person}{Gustavo Alonso}, {and} \bibinfo{person}{Donald Kossmann}.} \bibinfo{year}{2014}\natexlab{}.
\newblock \showarticletitle{Shared workload optimization}.
\newblock \bibinfo{journal}{\emph{Proceedings of the VLDB Endowment}} \bibinfo{volume}{7}, \bibinfo{number}{6}, \bibinfo{pages}{429--440}.
\newblock


\bibitem[Goodfellow et~al\mbox{.}(2016)]%
        {Goodfellow-et-al-2016}
\bibfield{author}{\bibinfo{person}{Ian Goodfellow}, \bibinfo{person}{Yoshua Bengio}, {and} \bibinfo{person}{Aaron Courville}.} \bibinfo{year}{2016}\natexlab{}.
\newblock \bibinfo{booktitle}{\emph{Deep Learning}}.
\newblock \bibinfo{publisher}{MIT Press}.
\newblock
\newblock
\shownote{\url{http://www.deeplearningbook.org}}.


\bibitem[Google(2022)]%
        {tfdataperformance}
\bibfield{author}{\bibinfo{person}{Google}.} \bibinfo{year}{2022}\natexlab{}.
\newblock \bibinfo{booktitle}{\emph{Better performance with the tf.data API}}.
\newblock
\urldef\tempurl%
\url{https://www.tensorflow.org/guide/data_performance}
\showURL{%
\tempurl}


\bibitem[{Google}(2022a)]%
        {google-cloud-pricing}
\bibfield{author}{\bibinfo{person}{{Google}}.} \bibinfo{year}{2022}\natexlab{a}.
\newblock \bibinfo{title}{{Google Cloud: All Pricing}}.
\newblock \bibinfo{howpublished}{\url{https://cloud.google.com/compute/all-pricing}}.
\newblock


\bibitem[{Google}(2022b)]%
        {google-cloud-tpu-config}
\bibfield{author}{\bibinfo{person}{{Google}}.} \bibinfo{year}{2022}\natexlab{b}.
\newblock \bibinfo{title}{{Google Cloud: TPU regions and zones}}.
\newblock \bibinfo{howpublished}{\url{https://cloud.google.com/tpu/docs/regions-zones}}.
\newblock


\bibitem[Google(2022)]%
        {tfdataserviceapidocumentation}
\bibfield{author}{\bibinfo{person}{Google}.} \bibinfo{year}{2022}\natexlab{}.
\newblock \bibinfo{booktitle}{\emph{tf.data service API documentation}}.
\newblock
\urldef\tempurl%
\url{https://www.tensorflow.org/api_docs/python/tf/data/experimental/service}
\showURL{%
\tempurl}


\bibitem[Google(2023)]%
        {colossus}
\bibfield{author}{\bibinfo{person}{Google}.} \bibinfo{year}{2023}\natexlab{}.
\newblock \bibinfo{booktitle}{\emph{{Colossus under the hood: a peek into Google’s scalable storage system}}}.
\newblock


\bibitem[{Google}(2023)]%
        {google-cloud-storage}
\bibfield{author}{\bibinfo{person}{{Google}}.} \bibinfo{year}{2023}\natexlab{}.
\newblock \bibinfo{title}{{Google Storage}}.
\newblock \bibinfo{howpublished}{\url{https://cloud.google.com/storage}}.
\newblock


\bibitem[Google(2023a)]%
        {grpc-documentation}
\bibfield{author}{\bibinfo{person}{Google}.} \bibinfo{year}{2023}\natexlab{a}.
\newblock \bibinfo{booktitle}{\emph{{gRPC Documentation}}}.
\newblock


\bibitem[Google(2023b)]%
        {gcp-network-pricing}
\bibfield{author}{\bibinfo{person}{Google}.} \bibinfo{year}{2023}\natexlab{b}.
\newblock \bibinfo{title}{{Network Pricing}}.
\newblock \bibinfo{howpublished}{\url{https://cloud.google.com/vpc/network-pricing}}.
\newblock


\bibitem[Graur et~al\mbox{.}(2022)]%
        {graur2022cachew}
\bibfield{author}{\bibinfo{person}{Dan Graur}, \bibinfo{person}{Damien Aymon}, \bibinfo{person}{Dan Kluser}, \bibinfo{person}{Tanguy Albrici}, \bibinfo{person}{Chandramohan~A Thekkath}, {and} \bibinfo{person}{Ana Klimovic}.} \bibinfo{year}{2022}\natexlab{}.
\newblock \showarticletitle{Cachew: Machine Learning Input Data Processing as a Service}. In \bibinfo{booktitle}{\emph{Proc. of USENIX ATC}}.
\newblock


\bibitem[Guirao et~al\mbox{.}(2019)]%
        {nvidia-dali}
\bibfield{author}{\bibinfo{person}{Joaquin~Anton Guirao}, \bibinfo{person}{Krzysztof Łęcki}, \bibinfo{person}{Janusz Lisiecki}, \bibinfo{person}{Serge Panev}, \bibinfo{person}{Michał Szołucha}, \bibinfo{person}{Albert Wolant}, {and} \bibinfo{person}{Michał Zientkiewicz}.} \bibinfo{year}{2019}\natexlab{}.
\newblock \bibinfo{title}{{Fast AI Data Preprocessing with NVIDIA DALI}}.
\newblock \bibinfo{howpublished}{\url{https://devblogs.nvidia.com/fast-ai-data-preprocessing-with-nvidia-dali}}.
\newblock


\bibitem[Harizopoulos et~al\mbox{.}(2005)]%
        {harizopoulos2005qpipe}
\bibfield{author}{\bibinfo{person}{Stavros Harizopoulos}, \bibinfo{person}{Vladislav Shkapenyuk}, {and} \bibinfo{person}{Anastassia Ailamaki}.} \bibinfo{year}{2005}\natexlab{}.
\newblock \showarticletitle{Qpipe: A simultaneously pipelined relational query engine}. In \bibinfo{booktitle}{\emph{Proceedings of the 2005 ACM SIGMOD international conference on Management of data}}. \bibinfo{pages}{383--394}.
\newblock


\bibitem[He et~al\mbox{.}(2016)]%
        {resnet}
\bibfield{author}{\bibinfo{person}{Kaiming He}, \bibinfo{person}{Xiangyu Zhang}, \bibinfo{person}{Shaoqing Ren}, {and} \bibinfo{person}{Jian Sun}.} \bibinfo{year}{2016}\natexlab{}.
\newblock \showarticletitle{Deep Residual Learning for Image Recognition}. In \bibinfo{booktitle}{\emph{Proc. of CVPR}}. \bibinfo{publisher}{{IEEE} Computer Society}.
\newblock
\urldef\tempurl%
\url{https://doi.org/10.1109/CVPR.2016.90}
\showDOI{\tempurl}


\bibitem[Hirzel et~al\mbox{.}(2014)]%
        {hirzel2014catalog}
\bibfield{author}{\bibinfo{person}{Martin Hirzel}, \bibinfo{person}{Robert Soul{\'e}}, \bibinfo{person}{Scott Schneider}, \bibinfo{person}{Bu{\u{g}}ra Gedik}, {and} \bibinfo{person}{Robert Grimm}.} \bibinfo{year}{2014}\natexlab{}.
\newblock \showarticletitle{A catalog of stream processing optimizations}.
\newblock \bibinfo{journal}{\emph{ACM Computing Surveys (CSUR)}} \bibinfo{volume}{46}, \bibinfo{number}{4} (\bibinfo{year}{2014}), \bibinfo{pages}{1--34}.
\newblock


\bibitem[HPA(2023)]%
        {kubernetes-hpa}
\bibfield{author}{\bibinfo{person}{Kubernetes HPA}.} \bibinfo{year}{2023}\natexlab{}.
\newblock \bibinfo{booktitle}{\emph{{Kubernetes Horizontal Pod Autoscaler Documentation}}}.
\newblock
\urldef\tempurl%
\url{https://kubernetes.io/docs/tasks/run-application/horizontal-pod-autoscale/}
\showURL{%
\tempurl}


\bibitem[Huyen(2022)]%
        {dmlsbook2022}
\bibfield{author}{\bibinfo{person}{Chip Huyen}.} \bibinfo{year}{2022}\natexlab{}.
\newblock \bibinfo{booktitle}{\emph{{Designing Machine Learning Systems}}}.
\newblock \bibinfo{publisher}{O'Reilly Media}, \bibinfo{address}{USA}.
\newblock
\showISBNx{978-1801819312}


\bibitem[Jiang et~al\mbox{.}(2020)]%
        {BytePS}
\bibfield{author}{\bibinfo{person}{Yimin Jiang}, \bibinfo{person}{Yibo Zhu}, \bibinfo{person}{Chang Lan}, \bibinfo{person}{Bairen Yi}, \bibinfo{person}{Yong Cui}, {and} \bibinfo{person}{Chuanxiong Guo}.} \bibinfo{year}{2020}\natexlab{}.
\newblock \showarticletitle{A Unified Architecture for Accelerating Distributed {DNN} Training in Heterogeneous {GPU/CPU} Clusters}. In \bibinfo{booktitle}{\emph{14th USENIX Symposium on Operating Systems Design and Implementation (OSDI 20)}}.
\newblock


\bibitem[Jouppi et~al\mbox{.}(2023)]%
        {jouppi2023tpu}
\bibfield{author}{\bibinfo{person}{Norman~P Jouppi}, \bibinfo{person}{George Kurian}, \bibinfo{person}{Sheng Li}, \bibinfo{person}{Peter Ma}, \bibinfo{person}{Rahul Nagarajan}, \bibinfo{person}{Lifeng Nai}, \bibinfo{person}{Nishant Patil}, \bibinfo{person}{Suvinay Subramanian}, \bibinfo{person}{Andy Swing}, \bibinfo{person}{Brian Towles}, {et~al\mbox{.}}} \bibinfo{year}{2023}\natexlab{}.
\newblock \showarticletitle{Tpu v4: An optically reconfigurable supercomputer for machine learning with hardware support for embeddings}.
\newblock \bibinfo{journal}{\emph{arXiv preprint arXiv:2304.01433}} (\bibinfo{year}{2023}).
\newblock


\bibitem[Jouppi et~al\mbox{.}(2020)]%
        {tpuv2v3-cacm}
\bibfield{author}{\bibinfo{person}{Norman~P. Jouppi}, \bibinfo{person}{Doe~Hyun Yoon}, \bibinfo{person}{George Kurian}, \bibinfo{person}{Sheng Li}, \bibinfo{person}{Nishant Patil}, \bibinfo{person}{James Laudon}, \bibinfo{person}{Cliff Young}, {and} \bibinfo{person}{David Patterson}.} \bibinfo{year}{2020}\natexlab{}.
\newblock \showarticletitle{A Domain-Specific Supercomputer for Training Deep Neural Networks}.
\newblock \bibinfo{journal}{\emph{Commun. ACM}} \bibinfo{volume}{63}, \bibinfo{number}{7} (\bibinfo{year}{2020}).
\newblock


\bibitem[Jouppi et~al\mbox{.}(2017)]%
        {tpuv1}
\bibfield{author}{\bibinfo{person}{Norman~P. Jouppi}, \bibinfo{person}{Cliff Young}, \bibinfo{person}{Nishant Patil}, \bibinfo{person}{David Patterson}, \bibinfo{person}{Gaurav Agrawal}, \bibinfo{person}{Raminder Bajwa}, \bibinfo{person}{Sarah Bates}, \bibinfo{person}{Suresh Bhatia}, \bibinfo{person}{Nan Boden}, \bibinfo{person}{Al Borchers}, \bibinfo{person}{Rick Boyle}, \bibinfo{person}{Pierre-luc Cantin}, \bibinfo{person}{Clifford Chao}, \bibinfo{person}{Chris Clark}, \bibinfo{person}{Jeremy Coriell}, \bibinfo{person}{Mike Daley}, \bibinfo{person}{Matt Dau}, \bibinfo{person}{Jeffrey Dean}, \bibinfo{person}{Ben Gelb}, \bibinfo{person}{Tara~Vazir Ghaemmaghami}, \bibinfo{person}{Rajendra Gottipati}, \bibinfo{person}{William Gulland}, \bibinfo{person}{Robert Hagmann}, \bibinfo{person}{C.~Richard Ho}, \bibinfo{person}{Doug Hogberg}, \bibinfo{person}{John Hu}, \bibinfo{person}{Robert Hundt}, \bibinfo{person}{Dan Hurt}, \bibinfo{person}{Julian Ibarz}, \bibinfo{person}{Aaron Jaffey}, \bibinfo{person}{Alek
  Jaworski}, \bibinfo{person}{Alexander Kaplan}, \bibinfo{person}{Harshit Khaitan}, \bibinfo{person}{Daniel Killebrew}, \bibinfo{person}{Andy Koch}, \bibinfo{person}{Naveen Kumar}, \bibinfo{person}{Steve Lacy}, \bibinfo{person}{James Laudon}, \bibinfo{person}{James Law}, \bibinfo{person}{Diemthu Le}, \bibinfo{person}{Chris Leary}, \bibinfo{person}{Zhuyuan Liu}, \bibinfo{person}{Kyle Lucke}, \bibinfo{person}{Alan Lundin}, \bibinfo{person}{Gordon MacKean}, \bibinfo{person}{Adriana Maggiore}, \bibinfo{person}{Maire Mahony}, \bibinfo{person}{Kieran Miller}, \bibinfo{person}{Rahul Nagarajan}, \bibinfo{person}{Ravi Narayanaswami}, \bibinfo{person}{Ray Ni}, \bibinfo{person}{Kathy Nix}, \bibinfo{person}{Thomas Norrie}, \bibinfo{person}{Mark Omernick}, \bibinfo{person}{Narayana Penukonda}, \bibinfo{person}{Andy Phelps}, \bibinfo{person}{Jonathan Ross}, \bibinfo{person}{Matt Ross}, \bibinfo{person}{Amir Salek}, \bibinfo{person}{Emad Samadiani}, \bibinfo{person}{Chris Severn}, \bibinfo{person}{Gregory Sizikov},
  \bibinfo{person}{Matthew Snelham}, \bibinfo{person}{Jed Souter}, \bibinfo{person}{Dan Steinberg}, \bibinfo{person}{Andy Swing}, \bibinfo{person}{Mercedes Tan}, \bibinfo{person}{Gregory Thorson}, \bibinfo{person}{Bo Tian}, \bibinfo{person}{Horia Toma}, \bibinfo{person}{Erick Tuttle}, \bibinfo{person}{Vijay Vasudevan}, \bibinfo{person}{Richard Walter}, \bibinfo{person}{Walter Wang}, \bibinfo{person}{Eric Wilcox}, {and} \bibinfo{person}{Doe~Hyun Yoon}.} \bibinfo{year}{2017}\natexlab{}.
\newblock \showarticletitle{In-Datacenter Performance Analysis of a Tensor Processing Unit}. In \bibinfo{booktitle}{\emph{Proc. of ISCA}} (Toronto, ON, Canada) \emph{(\bibinfo{series}{ISCA ’17})}. \bibinfo{publisher}{Association for Computing Machinery}, \bibinfo{address}{New York, NY, USA}.
\newblock
\urldef\tempurl%
\url{https://doi.org/10.1145/3079856.3080246}
\showDOI{\tempurl}


\bibitem[Kakaraparthy et~al\mbox{.}(2019)]%
        {oneaccess}
\bibfield{author}{\bibinfo{person}{Aarati Kakaraparthy}, \bibinfo{person}{Abhay Venkatesh}, \bibinfo{person}{Amar Phanishayee}, {and} \bibinfo{person}{Shivaram Venkataraman}.} \bibinfo{year}{2019}\natexlab{}.
\newblock \showarticletitle{The Case for Unifying Data Loading in Machine Learning Clusters}. In \bibinfo{booktitle}{\emph{11th {USENIX} Workshop on Hot Topics in Cloud Computing (HotCloud 19)}}.
\newblock


\bibitem[Klimovic et~al\mbox{.}(2016)]%
        {disagg-flash}
\bibfield{author}{\bibinfo{person}{Ana Klimovic}, \bibinfo{person}{Christos Kozyrakis}, \bibinfo{person}{Eno Thereska}, \bibinfo{person}{Binu John}, {and} \bibinfo{person}{Sanjeev Kumar}.} \bibinfo{year}{2016}\natexlab{}.
\newblock \showarticletitle{Flash Storage Disaggregation}. In \bibinfo{booktitle}{\emph{Proc. EuroSys}} \emph{(\bibinfo{series}{EuroSys '16})}. Article \bibinfo{articleno}{29}.
\newblock


\bibitem[Kubernetes(2023)]%
        {kubernetes}
\bibfield{author}{\bibinfo{person}{Kubernetes}.} \bibinfo{year}{2023}\natexlab{}.
\newblock \bibinfo{booktitle}{\emph{{kubernetes Documentation}}}.
\newblock
\urldef\tempurl%
\url{https://kubernetes.io/docs/home/}
\showURL{%
\tempurl}


\bibitem[Kuchnik et~al\mbox{.}(2022)]%
        {plumber}
\bibfield{author}{\bibinfo{person}{Michael Kuchnik}, \bibinfo{person}{Ana Klimovic}, \bibinfo{person}{Jiri Simsa}, \bibinfo{person}{Virginia Smith}, {and} \bibinfo{person}{George Amvrosiadis}.} \bibinfo{year}{2022}\natexlab{}.
\newblock \showarticletitle{Plumber: Diagnosing and Removing Performance Bottlenecks in Machine Learning Data Pipelines}. In \bibinfo{booktitle}{\emph{Proc. of Machine Learning and Systems}}, Vol.~\bibinfo{volume}{4}. \bibinfo{pages}{33--51}.
\newblock


\bibitem[Kumar and Sivathanu(2020)]%
        {quiver}
\bibfield{author}{\bibinfo{person}{Abhishek~Vijaya Kumar} {and} \bibinfo{person}{Muthian Sivathanu}.} \bibinfo{year}{2020}\natexlab{}.
\newblock \showarticletitle{Quiver: An Informed Storage Cache for Deep Learning}. In \bibinfo{booktitle}{\emph{Proc. of FAST}}.
\newblock


\bibitem[Lee et~al\mbox{.}(2021)]%
        {revamper}
\bibfield{author}{\bibinfo{person}{Gyewon Lee}, \bibinfo{person}{Irene Lee}, \bibinfo{person}{Hyeonmin Ha}, \bibinfo{person}{Kyunggeun Lee}, \bibinfo{person}{Hwarim Hyun}, \bibinfo{person}{Ahnjae Shin}, {and} \bibinfo{person}{Byung-Gon Chun}.} \bibinfo{year}{2021}\natexlab{}.
\newblock \showarticletitle{Refurbish Your Training Data: Reusing Partially Augmented Samples for Faster Deep Neural Network Training}. In \bibinfo{booktitle}{\emph{Proc. of USENIX ATC}}.
\newblock


\bibitem[Li et~al\mbox{.}(2014)]%
        {param-server}
\bibfield{author}{\bibinfo{person}{Mu Li}, \bibinfo{person}{David~G. Andersen}, \bibinfo{person}{Jun~Woo Park}, \bibinfo{person}{Alexander~J. Smola}, \bibinfo{person}{Amr Ahmed}, \bibinfo{person}{Vanja Josifovski}, \bibinfo{person}{James Long}, \bibinfo{person}{Eugene~J. Shekita}, {and} \bibinfo{person}{Bor-Yiing Su}.} \bibinfo{year}{2014}\natexlab{}.
\newblock \showarticletitle{Scaling Distributed Machine Learning with the Parameter Server}. In \bibinfo{booktitle}{\emph{Proceedings of the 11th USENIX Conference on Operating Systems Design and Implementation}} \emph{(\bibinfo{series}{OSDI'14})}.
\newblock


\bibitem[Lin et~al\mbox{.}(2017)]%
        {lin2017focal}
\bibfield{author}{\bibinfo{person}{Tsung-Yi Lin}, \bibinfo{person}{Priya Goyal}, \bibinfo{person}{Ross Girshick}, \bibinfo{person}{Kaiming He}, {and} \bibinfo{person}{Piotr Doll{\'a}r}.} \bibinfo{year}{2017}\natexlab{}.
\newblock \showarticletitle{Focal loss for dense object detection}. In \bibinfo{booktitle}{\emph{Proceedings of the IEEE international conference on computer vision}}. \bibinfo{pages}{2980--2988}.
\newblock


\bibitem[Lin et~al\mbox{.}(2014)]%
        {coco}
\bibfield{author}{\bibinfo{person}{Tsung-Yi Lin}, \bibinfo{person}{Michael Maire}, \bibinfo{person}{Serge Belongie}, \bibinfo{person}{James Hays}, \bibinfo{person}{Pietro Perona}, \bibinfo{person}{Deva Ramanan}, \bibinfo{person}{Piotr Dollár}, {and} \bibinfo{person}{C.~Lawrence Zitnick}.} \bibinfo{year}{2014}\natexlab{}.
\newblock \showarticletitle{Microsoft COCO: Common Objects in Context}. In \bibinfo{booktitle}{\emph{Proc. of ECCV}} (2014-01-01). \bibinfo{address}{Zürich}.
\newblock
\urldef\tempurl%
\url{/se3/wp-content/uploads/2014/09/coco_eccv.pdf, http://mscoco.org}
\showURL{%
\tempurl}
\newblock
\shownote{Oral}.


\bibitem[Marroquin et~al\mbox{.}(2018)]%
        {marroquin2018pay}
\bibfield{author}{\bibinfo{person}{Renato Marroquin}, \bibinfo{person}{Ingo M{\"u}ller}, \bibinfo{person}{Darko Makreshanski}, {and} \bibinfo{person}{Gustavo Alonso}.} \bibinfo{year}{2018}\natexlab{}.
\newblock \showarticletitle{Pay one, get hundreds for free: Reducing cloud costs through shared query execution}. In \bibinfo{booktitle}{\emph{Proceedings of the ACM Symposium on Cloud Computing}}. \bibinfo{pages}{439--450}.
\newblock


\bibitem[{Meta}(2022)]%
        {dpp-blog}
\bibfield{author}{\bibinfo{person}{{Meta}}.} \bibinfo{year}{2022}\natexlab{}.
\newblock \bibinfo{title}{{Scaling data ingestion for machine learning training at Meta}}.
\newblock \bibinfo{howpublished}{\url{https://engineering.fb.com/2022/09/19/ml-applications/data-ingestion-machine-learning-training-meta/}}.
\newblock


\bibitem[Min et~al\mbox{.}(2021)]%
        {smartNICjbofs}
\bibfield{author}{\bibinfo{person}{Jaehong Min}, \bibinfo{person}{Ming Liu}, \bibinfo{person}{Tapan Chugh}, \bibinfo{person}{Chenxingyu Zhao}, \bibinfo{person}{Andrew Wei}, \bibinfo{person}{In~Hwan Doh}, {and} \bibinfo{person}{Arvind Krishnamurthy}.} \bibinfo{year}{2021}\natexlab{}.
\newblock \showarticletitle{Gimbal: Enabling Multi-Tenant Storage Disaggregation on SmartNIC JBOFs}. In \bibinfo{booktitle}{\emph{Proc. of ACM SIGCOMM}} \emph{(\bibinfo{series}{SIGCOMM '21})}. \bibinfo{pages}{106–122}.
\newblock


\bibitem[MLCommons(2022)]%
        {mlperfconfig}
\bibfield{author}{\bibinfo{person}{MLCommons}.} \bibinfo{year}{2022}\natexlab{}.
\newblock \bibinfo{booktitle}{\emph{{ML Perf v2 Google Hardware Configurations}}}.
\newblock
\urldef\tempurl%
\url{https://github.com/mlcommons/training_results_v2.0/tree/main/Google/systems}
\showURL{%
\tempurl}


\bibitem[Mohan et~al\mbox{.}(2021a)]%
        {mohan2021checkfreq}
\bibfield{author}{\bibinfo{person}{Jayashree Mohan}, \bibinfo{person}{Amar Phanishayee}, {and} \bibinfo{person}{Vijay Chidambaram}.} \bibinfo{year}{2021}\natexlab{a}.
\newblock \showarticletitle{$\{$CheckFreq$\}$: Frequent,$\{$Fine-Grained$\}$$\{$DNN$\}$ Checkpointing}. In \bibinfo{booktitle}{\emph{19th USENIX Conference on File and Storage Technologies (FAST 21)}}. \bibinfo{pages}{203--216}.
\newblock


\bibitem[Mohan et~al\mbox{.}(2021b)]%
        {coordl}
\bibfield{author}{\bibinfo{person}{Jayashree Mohan}, \bibinfo{person}{Amar Phanishayee}, \bibinfo{person}{Ashish Raniwala}, {and} \bibinfo{person}{Vijay Chidambaram}.} \bibinfo{year}{2021}\natexlab{b}.
\newblock \bibinfo{title}{Analyzing and Mitigating Data Stalls in DNN Training}.
\newblock
\newblock
\showeprint[arxiv]{2007.06775}~[cs.DC]


\bibitem[Murray et~al\mbox{.}(2021)]%
        {murray2021tfdata}
\bibfield{author}{\bibinfo{person}{Derek~G. Murray}, \bibinfo{person}{Ji\v{r}\'{i} \v{S}im\v{s}a}, \bibinfo{person}{Ana Klimovic}, {and} \bibinfo{person}{Ihor Indyk}.} \bibinfo{year}{2021}\natexlab{}.
\newblock \showarticletitle{tf.data: A Machine Learning Data Processing Framework}.
\newblock \bibinfo{journal}{\emph{Proc. VLDB Endow.}} \bibinfo{volume}{14}, \bibinfo{number}{12} (\bibinfo{year}{2021}).
\newblock


\bibitem[{MXNET}(2018)]%
        {mxnet_dataio}
\bibfield{author}{\bibinfo{person}{{MXNET}}.} \bibinfo{year}{2018}\natexlab{}.
\newblock \bibinfo{title}{{Designing Efficient Data Loaders for Deep Learning}}.
\newblock \bibinfo{howpublished}{\url{https://mxnet.apache.org/api/architecture/note_data_loading}}.
\newblock


\bibitem[Paszke et~al\mbox{.}(2019)]%
        {pytorch}
\bibfield{author}{\bibinfo{person}{Adam Paszke}, \bibinfo{person}{Sam Gross}, \bibinfo{person}{Francisco Massa}, \bibinfo{person}{Adam Lerer}, \bibinfo{person}{James Bradbury}, \bibinfo{person}{Gregory Chanan}, \bibinfo{person}{Trevor Killeen}, \bibinfo{person}{Zeming Lin}, \bibinfo{person}{Natalia Gimelshein}, \bibinfo{person}{Luca Antiga}, \bibinfo{person}{Alban Desmaison}, \bibinfo{person}{Andreas Kopf}, \bibinfo{person}{Edward Yang}, \bibinfo{person}{Zachary DeVito}, \bibinfo{person}{Martin Raison}, \bibinfo{person}{Alykhan Tejani}, \bibinfo{person}{Sasank Chilamkurthy}, \bibinfo{person}{Benoit Steiner}, \bibinfo{person}{Lu Fang}, \bibinfo{person}{Junjie Bai}, {and} \bibinfo{person}{Soumith Chintala}.} \bibinfo{year}{2019}\natexlab{}.
\newblock \showarticletitle{{PyTorch: An Imperative Style, High-Performance Deep Learning Library}}.
\newblock In \bibinfo{booktitle}{\emph{Advances in Neural Information Processing Systems 32}}. \bibinfo{publisher}{Curran Associates, Inc.}, \bibinfo{pages}{8024--8035}.
\newblock
\urldef\tempurl%
\url{http://papers.neurips.cc/paper/9015-pytorch-an-imperative-style-high-performance-deep-learning-library.pdf}
\showURL{%
\tempurl}


\bibitem[Rzadca et~al\mbox{.}(2020)]%
        {rzadca2020autopilot}
\bibfield{author}{\bibinfo{person}{Krzysztof Rzadca}, \bibinfo{person}{Pawel Findeisen}, \bibinfo{person}{Jacek Swiderski}, \bibinfo{person}{Przemyslaw Zych}, \bibinfo{person}{Przemyslaw Broniek}, \bibinfo{person}{Jarek Kusmierek}, \bibinfo{person}{Pawel Nowak}, \bibinfo{person}{Beata Strack}, \bibinfo{person}{Piotr Witusowski}, \bibinfo{person}{Steven Hand}, {et~al\mbox{.}}} \bibinfo{year}{2020}\natexlab{}.
\newblock \showarticletitle{Autopilot: workload autoscaling at google}. In \bibinfo{booktitle}{\emph{Proc. of the Fifteenth European Conference on Computer Systems}}.
\newblock


\bibitem[Shan et~al\mbox{.}(2018)]%
        {legoos}
\bibfield{author}{\bibinfo{person}{Yizhou Shan}, \bibinfo{person}{Yutong Huang}, \bibinfo{person}{Yilun Chen}, {and} \bibinfo{person}{Yiying Zhang}.} \bibinfo{year}{2018}\natexlab{}.
\newblock \showarticletitle{{LegoOS}: A Disseminated, Distributed {OS} for Hardware Resource Disaggregation}. In \bibinfo{booktitle}{\emph{Proc. of OSDI}}.
\newblock


\bibitem[Shenoy and Ozsoyoglu(1987)]%
        {semantic_query_opt}
\bibfield{author}{\bibinfo{person}{Sreekumar~T. Shenoy} {and} \bibinfo{person}{Z.~Meral Ozsoyoglu}.} \bibinfo{year}{1987}\natexlab{}.
\newblock \showarticletitle{A System for Semantic Query Optimization}. In \bibinfo{booktitle}{\emph{Proceedings of the 1987 ACM SIGMOD International Conference on Management of Data}} (San Francisco, California, USA) \emph{(\bibinfo{series}{SIGMOD '87})}. \bibinfo{publisher}{Association for Computing Machinery}, \bibinfo{address}{New York, NY, USA}, \bibinfo{pages}{181–195}.
\newblock
\showISBNx{0897912365}
\urldef\tempurl%
\url{https://doi.org/10.1145/38713.38736}
\showDOI{\tempurl}


\bibitem[Shorten and Khoshgoftaar(2019)]%
        {shorten2019survey}
\bibfield{author}{\bibinfo{person}{Connor Shorten} {and} \bibinfo{person}{Taghi~M Khoshgoftaar}.} \bibinfo{year}{2019}\natexlab{}.
\newblock \showarticletitle{A survey on image data augmentation for deep learning}.
\newblock \bibinfo{journal}{\emph{Journal of big data}} \bibinfo{volume}{6}, \bibinfo{number}{1} (\bibinfo{year}{2019}), \bibinfo{pages}{1--48}.
\newblock


\bibitem[Shorten et~al\mbox{.}(2021)]%
        {shorten2021text}
\bibfield{author}{\bibinfo{person}{Connor Shorten}, \bibinfo{person}{Taghi~M Khoshgoftaar}, {and} \bibinfo{person}{Borko Furht}.} \bibinfo{year}{2021}\natexlab{}.
\newblock \showarticletitle{Text data augmentation for deep learning}.
\newblock \bibinfo{journal}{\emph{Journal of big Data}}  \bibinfo{volume}{8} (\bibinfo{year}{2021}), \bibinfo{pages}{1--34}.
\newblock


\bibitem[Simard et~al\mbox{.}(2003)]%
        {best-practices-cnns}
\bibfield{author}{\bibinfo{person}{Patrice~Y. Simard}, \bibinfo{person}{Dave Steinkraus}, {and} \bibinfo{person}{John~C. Platt}.} \bibinfo{year}{2003}\natexlab{}.
\newblock \showarticletitle{Best Practices for Convolutional Neural Networks Applied to Visual Document Analysis}. In \bibinfo{booktitle}{\emph{Proc. of ICDAR}} \emph{(\bibinfo{series}{ICDAR ’03})}. \bibinfo{publisher}{IEEE Computer Society}, \bibinfo{address}{USA}, \bibinfo{numpages}{1}~pages.
\newblock
\showISBNx{0769519601}


\bibitem[Spark(2023)]%
        {spark-streaming-performance}
\bibfield{author}{\bibinfo{person}{Apache Spark}.} \bibinfo{year}{2023}\natexlab{}.
\newblock \bibinfo{title}{{Spark Streaming Programming Guide}}.
\newblock \bibinfo{howpublished}{\url{https://spark.apache.org/docs/latest/streaming-programming-guide.html}}.
\newblock


\bibitem[{TensorFlow}(2022a)]%
        {tfdataservice-doc}
\bibfield{author}{\bibinfo{person}{{TensorFlow}}.} \bibinfo{year}{2022}\natexlab{a}.
\newblock \bibinfo{title}{{Module: tf.data.experimental.service}}.
\newblock \bibinfo{howpublished}{\url{https://www.tensorflow.org/api_docs/python/tf/data/experimental/service}}.
\newblock


\bibitem[{TensorFlow}(2022b)]%
        {tfdata}
\bibfield{author}{\bibinfo{person}{{TensorFlow}}.} \bibinfo{year}{2022}\natexlab{b}.
\newblock \bibinfo{title}{{tf.data: Build TensorFlow input pipelines}}.
\newblock \bibinfo{howpublished}{\url{https://www.tensorflow.org/guide/data}}.
\newblock


\bibitem[{TensorFlow}(2023a)]%
        {tensorflow-github}
\bibfield{author}{\bibinfo{person}{{TensorFlow}}.} \bibinfo{year}{2023}\natexlab{a}.
\newblock \bibinfo{title}{{Tensorflow}}.
\newblock \bibinfo{howpublished}{\url{https://github.com/tensorflow/tensorflow}}.
\newblock


\bibitem[{TensorFlow}(2023b)]%
        {modelgarden}
\bibfield{author}{\bibinfo{person}{{TensorFlow}}.} \bibinfo{year}{2023}\natexlab{b}.
\newblock \bibinfo{title}{{TensorFlow Model Garden}}.
\newblock \bibinfo{howpublished}{\url{https://github.com/tensorflow/models}}.
\newblock


\bibitem[Tirmazi et~al\mbox{.}(2020)]%
        {tirmazi2020borg}
\bibfield{author}{\bibinfo{person}{Muhammad Tirmazi}, \bibinfo{person}{Adam Barker}, \bibinfo{person}{Nan Deng}, \bibinfo{person}{Md~E Haque}, \bibinfo{person}{Zhijing~Gene Qin}, \bibinfo{person}{Steven Hand}, \bibinfo{person}{Mor Harchol-Balter}, {and} \bibinfo{person}{John Wilkes}.} \bibinfo{year}{2020}\natexlab{}.
\newblock \showarticletitle{Borg: the next generation}. In \bibinfo{booktitle}{\emph{Proceedings of the fifteenth European conference on computer systems}}. \bibinfo{pages}{1--14}.
\newblock


\bibitem[Um et~al\mbox{.}(2023)]%
        {fastflow}
\bibfield{author}{\bibinfo{person}{Taegeon Um}, \bibinfo{person}{Byungsoo Oh}, \bibinfo{person}{Byeongchan Seo}, \bibinfo{person}{Minhyeok Kweun}, \bibinfo{person}{Goeun Kim}, {and} \bibinfo{person}{Woo-Yeon Lee}.} \bibinfo{year}{2023}\natexlab{}.
\newblock \showarticletitle{{FastFlow}: Accelerating Deep Learning Model Training with Smart Offloading of Input Data Pipeline}.
\newblock \bibinfo{journal}{\emph{Proceedings of the VLDB Endowment}} \bibinfo{volume}{16}, \bibinfo{number}{5} (\bibinfo{year}{2023}), \bibinfo{pages}{1086--1099}.
\newblock


\bibitem[Verbraeken et~al\mbox{.}(2020)]%
        {distrib-ML}
\bibfield{author}{\bibinfo{person}{Joost Verbraeken}, \bibinfo{person}{Matthijs Wolting}, \bibinfo{person}{Jonathan Katzy}, \bibinfo{person}{Jeroen Kloppenburg}, \bibinfo{person}{Tim Verbelen}, {and} \bibinfo{person}{Jan~S. Rellermeyer}.} \bibinfo{year}{2020}\natexlab{}.
\newblock \showarticletitle{A Survey on Distributed Machine Learning}.
\newblock \bibinfo{journal}{\emph{ACM Comput. Surv.}} \bibinfo{volume}{53}, \bibinfo{number}{2}, Article \bibinfo{articleno}{30} (\bibinfo{date}{mar} \bibinfo{year}{2020}).
\newblock


\bibitem[Verma et~al\mbox{.}(2015)]%
        {verma2015borg}
\bibfield{author}{\bibinfo{person}{Abhishek Verma}, \bibinfo{person}{Luis Pedrosa}, \bibinfo{person}{Madhukar Korupolu}, \bibinfo{person}{David Oppenheimer}, \bibinfo{person}{Eric Tune}, {and} \bibinfo{person}{John Wilkes}.} \bibinfo{year}{2015}\natexlab{}.
\newblock \showarticletitle{Large-scale cluster management at Google with Borg}. In \bibinfo{booktitle}{\emph{Proc. of EuroSys}}.
\newblock


\bibitem[VPA(2023)]%
        {kubernetes-vpa}
\bibfield{author}{\bibinfo{person}{Kubernetes VPA}.} \bibinfo{year}{2023}\natexlab{}.
\newblock \bibinfo{booktitle}{\emph{{Kubernetes Vertical Pod Autoscaler Documentation}}}.
\newblock
\urldef\tempurl%
\url{https://cloud.google.com/kubernetes-engine/docs/concepts/verticalpodautoscaler}
\showURL{%
\tempurl}


\bibitem[Yan et~al\mbox{.}(2020)]%
        {neural-data-server}
\bibfield{author}{\bibinfo{person}{Xi Yan}, \bibinfo{person}{David Acuna}, {and} \bibinfo{person}{Sanja Fidler}.} \bibinfo{year}{2020}\natexlab{}.
\newblock \showarticletitle{Neural Data Server: A Large-Scale Search Engine for Transfer Learning Data}. In \bibinfo{booktitle}{\emph{Proc. of the IEEE/CVF Conference on Computer Vision and Pattern Recognition (CVPR)}}.
\newblock


\bibitem[Zaharia et~al\mbox{.}(2010)]%
        {spark}
\bibfield{author}{\bibinfo{person}{Matei Zaharia}, \bibinfo{person}{Mosharaf Chowdhury}, \bibinfo{person}{Michael~J. Franklin}, \bibinfo{person}{Scott Shenker}, {and} \bibinfo{person}{Ion Stoica}.} \bibinfo{year}{2010}\natexlab{}.
\newblock \showarticletitle{Spark: Cluster Computing with Working Sets}. In \bibinfo{booktitle}{\emph{Proc. of HotCloud}} (Boston, MA) \emph{(\bibinfo{series}{HotCloud’10})}. \bibinfo{publisher}{USENIX Association}, \bibinfo{address}{USA}, \bibinfo{pages}{10}.
\newblock


\bibitem[Zhao et~al\mbox{.}(2022)]%
        {dpp}
\bibfield{author}{\bibinfo{person}{Mark Zhao}, \bibinfo{person}{Niket Agarwal}, \bibinfo{person}{Aarti Basant}, \bibinfo{person}{Bu\u{g}ra Gedik}, \bibinfo{person}{Satadru Pan}, \bibinfo{person}{Mustafa Ozdal}, \bibinfo{person}{Rakesh Komuravelli}, \bibinfo{person}{Jerry Pan}, \bibinfo{person}{Tianshu Bao}, \bibinfo{person}{Haowei Lu}, \bibinfo{person}{Sundaram Narayanan}, \bibinfo{person}{Jack Langman}, \bibinfo{person}{Kevin Wilfong}, \bibinfo{person}{Harsha Rastogi}, \bibinfo{person}{Carole-Jean Wu}, \bibinfo{person}{Christos Kozyrakis}, {and} \bibinfo{person}{Parik Pol}.} \bibinfo{year}{2022}\natexlab{}.
\newblock \showarticletitle{Understanding Data Storage and Ingestion for Large-Scale Deep Recommendation Model Training: Industrial Product}. In \bibinfo{booktitle}{\emph{Proc. of ISCA}} \emph{(\bibinfo{series}{ISCA '22})}.
\newblock


\end{thebibliography}

\end{document}